\pdfoutput=1
\documentclass[11pt,a4paper]{article}

\usepackage[a4paper,margin=1in,headheight=28pt,headsep=12pt]{geometry}

\usepackage[utf8]{inputenc}
\usepackage[T1]{fontenc}

\usepackage[x11names]{xcolor}

\usepackage{amsmath, amssymb, mathtools, bm, relsize}

\usepackage{booktabs, multirow, tabularx, ragged2e, makecell}

\usepackage{graphicx}

\usepackage{enumitem}

\usepackage{listings}

\usepackage[font=small,labelfont=bf,labelsep=period]{caption}
\renewcommand{\figurename}{Fig.}

\usepackage{csquotes}

\usepackage{setspace}
\usepackage{xspace}

\usepackage{XCharter}
\usepackage[scaled=1.06]{zlmtt}
\usepackage[xcharter,bigdelims,vvarbb]{newtxmath}

\usepackage{fancyhdr}
\usepackage{titlesec, titling}
\usepackage{tocloft}

\usepackage{authblk}

\usepackage[
  backend=biber,
  style=nature,
  sorting=none,
  maxbibnames=3,
  natbib=true,
]{biblatex}

\usepackage[
  pagebackref=false,
  breaklinks=true,
  colorlinks=true,
  bookmarks=true,
  citecolor=Blue3,
  urlcolor=Blue3,
  linkcolor=Blue3,
]{hyperref}

\usepackage[capitalize]{cleveref}

\crefname{figure}{Fig.}{Figs.}
\Crefname{figure}{Fig.}{Figs.}
\crefname{section}{Section}{Sections}
\Crefname{section}{Section}{Sections}
\crefname{subsection}{Section}{Sections}
\Crefname{subsection}{Section}{Sections}

\newcommand{\snoteone}{\hyperref[snote:prompts]{Supplementary Note~1}}
\newcommand{\snotetwo}{\hyperref[snote:objectives]{Supplementary Note~2}}
\newcommand{\sappone}{\hyperref[sapp:proposal]{Supplementary Appendix~1}}

\newcommand{\sfigref}[1]{Supplementary Fig.~\labelcref{#1}}

\newcommand{\promptlisting}[2]{%
  \par\addvspace{1.2em}%
  \noindent
  \begin{minipage}{\linewidth}%
    \setlength{\parskip}{0pt}%
    \lstinputlisting[style=promptbox, title={\textbf{#1}},
      aboveskip=0pt, belowskip=0pt]{#2}%
  \end{minipage}%
  \par\addvspace{1.2em}%
}

\newcommand{\proposalpage}[3][\textheight]{%
  \begin{center}
    \includegraphics[width=\linewidth, height=#1, keepaspectratio,
      viewport=#3, clip]{#2}%
  \end{center}%
}

\definecolor{promptbg}{rgb}{0.976, 0.980, 0.988}
\definecolor{promptrule}{rgb}{0.70, 0.73, 0.78}

\lstdefinestyle{promptbox}{
  basicstyle=\ttfamily\scriptsize,
  breaklines=true,
  breakatwhitespace=false,
  columns=flexible,
  keepspaces=true,
  showstringspaces=false,
  extendedchars=true,
  frame=single,
  rulecolor=\color{promptrule},
  backgroundcolor=\color{promptbg},
  framesep=5pt,
  framexleftmargin=5pt,
  framexrightmargin=5pt,
  xleftmargin=5pt,
  xrightmargin=5pt,
  aboveskip=1.2em,
  belowskip=1.2em,
  abovecaptionskip=4pt,
  literate=
    {—}{{---}}3
    {–}{{--}}2
    {’}{{'}}1
    {‘}{{`}}1
    {“}{{``}}2
    {”}{{''}}2
    {β}{{$\beta$}}1
    {μ}{{$\mu$}}1
    {µ}{{$\mu$}}1
    {≥}{{$\ge$}}1
    {≤}{{$\le$}}1
    {×}{{$\times$}}1
    {′}{{$'$}}1
    {·}{{$\cdot$}}1
    {…}{{\ldots}}3
}

\newcommand{\asteromorphmark}{%
  \raisebox{-2pt}{%
    \includegraphics[height=24pt]{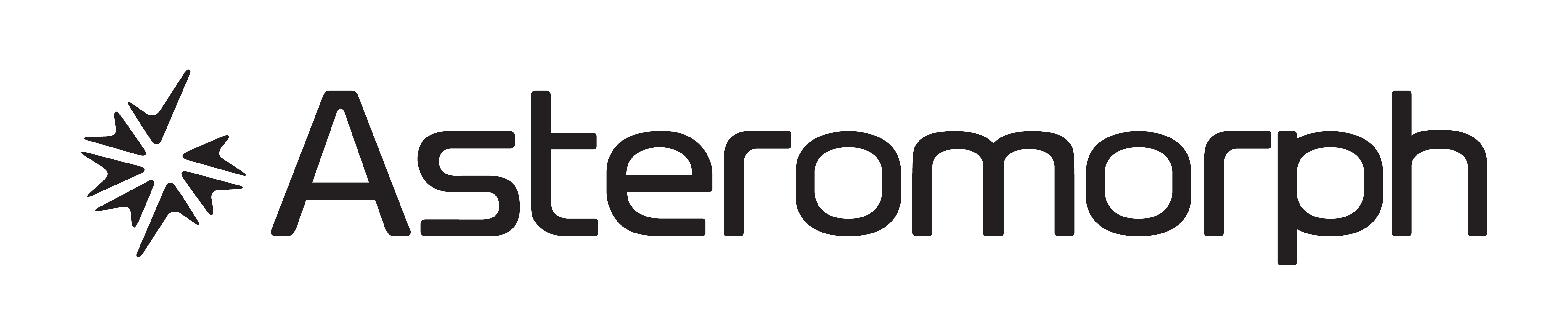}%
  }%
}

\fancypagestyle{titlepage}{%
  \fancyhf{}

  \fancyhead[L]{\asteromorphmark}
  \fancyhead[R]{\small\today}
}

\fancypagestyle{maincontent}{%
  \fancyhf{}

  \fancyfoot[C]{\small\thepage}
}

\titleformat{\section}
  {\normalfont\Large\bfseries\raggedright}
  {\thesection.}
  {1em}
  {}

\titlespacing*{\section}
  {0pt}
  {*3}
  {*1.5}

\titleformat{\subsection}
  {\normalfont\large\bfseries\raggedright}
  {\thesubsection}
  {1em}
  {}

\titlespacing*{\subsection}
  {0pt}
  {*2}
  {*1}

\begin{document}

% ==========================================================
% Title Page
% ==========================================================

\title{%
  \vspace{-0.4cm}%
  % Normal title size, set over two lines. The string measures 348.6pt at
  % 10pt against a 453.0pt text width, so 20pt breaks after "strategies"
  % and fills 97% of the first line. Do not go much above this: from about
  % 21pt the title costs an extra line of height and pushes the
  % equal-contribution/corresponding-author block off the title page.
  \fontsize{17}{20.4}\selectfont%
  \textbf{Structured AI ideation yields physical strategies for epithelial
  barrier repair}%
}

\renewcommand\Authands{, }
\renewcommand\Authfont{\normalsize}
\renewcommand\Affilfont{\small}

\author[1,*]{Geonho Nah}
\author[1,*]{Johyun Park}
\author[1]{Kyung-Ah Cho}
\author[1]{Suyoung Hwang}
\author[2]{YoonHee Lim}
\author[3,4]{Eun-Kyung Min}
\author[1]{Seunghyun Moon}
\author[1]{Donghyun Koh}
\author[1]{Seungwon Kim}
\author[1]{Jiho Park}
\author[1]{Hojin Yoo}
\author[1]{Myungsun Nam}
\author[1]{Haneul Choi}
\author[5,6,7]{Kyungsu Kim}
\author[3,4,8,\dag]{In-Sun Hong}
\author[2,4,8,\dag]{YunJae Jung}
\author[1,\dag]{Minhyeong Lee}

\affil[1]{%
  Asteromorph, Seoul 08788, Republic of Korea%
}

\affil[2]{%
  Department of Microbiology, College of Medicine,
  Gachon University, Incheon 21999, Republic of Korea%
}

\affil[3]{%
  Department of Molecular Medicine, College of Medicine,
  Gachon University, Incheon 21999, Republic of Korea%
}

\affil[4]{%
  Department of Health Science and Technology,
  Gachon Advanced Institute for Health Science \& Technology,
  Gachon University, Incheon 21999, Republic of Korea%
}

\affil[5]{%
  School of Transdisciplinary Innovations,
  Seoul National University, Seoul, Republic of Korea%
}

\affil[6]{%
  Department of Biomedical Sciences, College of Medicine,
  Seoul National University, Seoul, Republic of Korea%
}

\affil[7]{%
  Interdisciplinary Program in AI \& Graduate School of AI,
  Seoul National University, Seoul, Republic of Korea%
}

\affil[8]{%
  Lee Gil Ya Cancer and Diabetes Institute,
  Gachon University, Incheon 21999, Republic of Korea%
}

\date{}

\maketitle

\thispagestyle{titlepage}

\vspace{-1cm}

% Abstract and keywords
% ============== Abstract Section ============== %

\begin{abstract}
  \noindent
  AI systems increasingly support literature analysis, candidate evaluation and
  hypothesis refinement, but enabling researchers to explore research directions
  beyond established disciplinary approaches remains challenging. We developed
  Abelio, an AI system designed to generate novel research directions by
  expanding investigators' avenues of inquiry. Abelio reformulates research
  objectives, uses these alternative formulations to identify relevant elements
  outside conventional disciplinary boundaries, and refines the resulting
  hypotheses. Each stage integrates divergent exploration with convergent
  selection through a dedicated subsystem. In quantitative evaluations,
  proposals generated by the Ideation Core (Abelio's hypothesis-generation core)
  outperformed those of four frontier large language models in scientific
  soundness and hypothesis diversity, and three of four in distance from
  existing literature. Subsystem ablation showed that each subsystem contributes
  to the property it targets, with the contribution of Element Selection
  measurable before Refinement. To investigate intestinal epithelial barrier
  restoration, Abelio proposed physical interventions involving
  membrane-tethered poly(ethylene glycol) (PEG) and cholesterol-anchored
  complementary DNA. Both proposals were archived before experimentation and
  subsequently tested independently in damaged Caco-2 monolayers. PEG preserved
  barrier function in a membrane-surface-dependent manner, whereas complementary
  DNA enabled sequence-specific restoration that was reversible on demand. These
  findings demonstrate how structured research ideation can identify
  experimentally testable approaches beyond established disciplinary
  conventions.

  \medskip

\end{abstract}

\vfill

{\small
\noindent
\textsuperscript{*}These authors contributed equally to this work.

\noindent
\textsuperscript{\dag}Corresponding authors:
In-Sun Hong (\href{mailto:hongstem@gachon.ac.kr}{hongstem@gachon.ac.kr}),
YunJae Jung (\href{mailto:yjjung@gachon.ac.kr}{yjjung@gachon.ac.kr}),\\
Minhyeong Lee (\href{mailto:mh.lee@asteromorph.com}{mh.lee@asteromorph.com}).
}

% ==========================================================
% Table of Contents
% ==========================================================

\clearpage

{\hypersetup{linkcolor=black}
  \setlength{\parskip}{0pt}
  \tableofcontents
}

\thispagestyle{maincontent}

% ==========================================================
% Main Text
% ==========================================================

\clearpage
\pagestyle{maincontent}

% ============== Introduction Section ============== %

\section{Introduction}
\label{sec:introduction}

Artificial intelligence (AI) is increasingly recognized not merely as a
convenient tool but as an active participant in scientific research. Since the
breakthroughs achieved by
AlphaFold\autocite{senior2020improved,jumper2021highly,abramson2024accurate}, AI
systems have demonstrated remarkable capabilities across diverse scientific
tasks\autocite{degrave2022magnetic,lam2023learning,liu2023deep,merchant2023scaling,watson2023novo}.
Large language models (LLMs) have subsequently raised expectations toward
higher-order scientific reasoning, including hypothesis
generation\autocite{boiko2023autonomous,wang2023scientific,gao2024empowering,ifargan2025autonomous,ren2025towards,zhang2025exploring}.
Sakana AI's AI Scientist automates the entire machine learning research
pipeline\autocite{lu2026towards}. In mathematics, several AI systems have
autonomously solved open Erd\H{o}s problems with proofs formally verified using
the Lean proof assistant\autocite{sothanaphan2026resolution,tsoukalas2026advancing},
and generated Lean-verified results addressing long-standing open problems,
including one closely related to the Riemann
Hypothesis\autocite{openai2026ten,anthropic2026riemann,openai2026quasi}. In the
life sciences, Google DeepMind's Co-Scientist\autocite{gottweis2026accelerating}
and FutureHouse's Robin\autocite{ghareeb2026multi} have demonstrated
capabilities in target identification, drug repurposing, and mechanistic
explanation.

These advances demonstrate the capacity of AI systems to generate and refine
scientific hypotheses. Nevertheless, investigators must still determine how to
conceptualize a research objective and identify which directions warrant
exploration. Problem framing influences which knowledge is perceived as relevant
and which approaches are considered, potentially excluding alternative
possibilities. Generating additional hypotheses within an established framework
does not necessarily ensure consideration of fundamentally different approaches
to achieving the same objective. Reframing a research objective from multiple
scientific perspectives may uncover previously overlooked connections to
knowledge developed in other research contexts, thereby creating additional
opportunities for hypothesis generation.

Several advances in the life sciences illustrate how knowledge originating in
one research context can be applied to a different scientific problem, enabling
novel approaches. Optogenetics became feasible when a light-gated ion channel
from a green alga, previously characterized in photobiological research, was
expressed in neurons\autocite{nagel2003channelrhodopsin,boyden2005millisecond}.
Green fluorescent protein (GFP), now widely used in cell biology, was initially
isolated as a byproduct of studies on jellyfish bioluminescence and subsequently
adopted as a genetically encoded
marker\autocite{shimomura1962extraction,chalfie1994green}. Bioorthogonal
chemistry emerged through adaptation of a classical organic reaction, the
Staudinger reduction, for use in living systems\autocite{saxon2000cell}. These
examples demonstrate that the relevance of existing knowledge depends on how a
research problem is conceptualized, suggesting that alternative formulations of
a research objective may reveal connections to seemingly unrelated scientific
knowledge.

Here, we introduce Abelio, an AI system designed to expand the range of
approaches available to investigators. Given a research objective, Abelio
generates alternative formulations representing distinct scientific
perspectives. These formulations guide the identification of relevant scientific
elements, including those originating from fields not conventionally associated
with the problem, which are subsequently incorporated into hypotheses refined
through agent debate. Each stage integrates divergent exploration with
convergent selection through a dedicated subsystem, drawing on models of
creative reasoning to facilitate the exploration and evaluation of alternative
possibilities\autocite{cropley2006praise,dym2005engineering}. Across 50 research
objectives, proposals generated by the Ideation Core outperformed those of four
frontier LLMs in scientific soundness and three of four in terms of their
distance from the published literature. Ablation analysis revealed that
eliminating the problem formulation stage reduced hypothesis diversity,
supporting its contribution to expanding the range of approaches explored for a
given objective.

We subsequently investigated how an Abelio-generated research direction could be
developed into an experimentally testable intervention using a damaged
intestinal epithelial monolayer. The intestinal epithelium establishes a
selective barrier between the intestinal lumen and the internal environment,
facilitating nutrient absorption while limiting the passage of microorganisms,
toxins, and antigens\autocite{odenwald2017intestinal}. Barrier integrity depends
on tight junction complexes connecting adjacent epithelial
cells\autocite{moonwiriyakit2023tight,horowitz2023paracellular}, and its
disruption is common in chronic intestinal disorders, including inflammatory
bowel disease and celiac disease\autocite{schumann2017celiac}, which
collectively affect tens of millions of individuals
worldwide\autocite{kaplan2025global,singh2018global}. Aging, chronic
inflammation, infection, metabolic dysfunction, and drug exposure can also
increase paracellular permeability\autocite{camilleri2019leaky}. The
consequences extend beyond the intestine, as barrier dysfunction may facilitate
the systemic dissemination of gut-derived inflammatory
stimuli\autocite{odenwald2017intestinal,ahmad2017gut}. Existing research has
predominantly focused on regulating the expression and assembly of tight
junction
proteins\autocite{graham2019intracellular,schuppan2021randomized,ghosh2021regulation}.
However, this approach is limited by the underlying conditions contributing to
barrier dysfunction, as the capacity for epithelial repair is impaired during
aging and persistent low-grade
inflammation\autocite{pentinmikko2019notum,omrani2023ifngamma}, while few
interventions directly target the barrier
itself\autocite{moonwiriyakit2023tight,horowitz2023paracellular}.

In response to this objective, Abelio proposed physical interventions based on
polymer crowding and DNA-mediated membrane bridging. These strategies sought to
restore barrier function through physical interactions at cell surfaces,
extending beyond conventional approaches targeting tight junction protein
regulation. Neither strategy had previously been considered by the investigators
who defined the objective. We prospectively and independently evaluated both
approaches in an \textit{in vitro} barrier model, confirming the underlying
principles of both proposals, although certain aspects of the proposed designs
required modification. Collectively, the quantitative evaluations and
experimental findings demonstrate how structured research ideation can expand
the range of approaches considered for a scientific objective and generate
experimentally actionable starting points for further investigation.

% ============== Results: Abelio Overview Subsection ============== %

\section{Results}
\label{sec:results}

\subsection{Abelio overview}
\label{subsec:overview}

Abelio's architecture is founded on two principles. First, the reasoning
underlying the emergence of an unexplored approach should be explicitly
represented in the architecture and decomposed into formalized subprocesses.
Each agent therefore performs a clearly defined operation rather than responding
to an underspecified request for a novel hypothesis. Second, these operations
should follow iterative cycles of divergent expansion and convergent selection,
consistent with established models of creative
processes\autocite{cropley2006praise,dym2005engineering}. Each convergence
prioritizes preserving the breadth of the candidate space rather than
identifying a single optimum.

\begin{figure}[p]
  \centering
  \includegraphics[width=\linewidth]{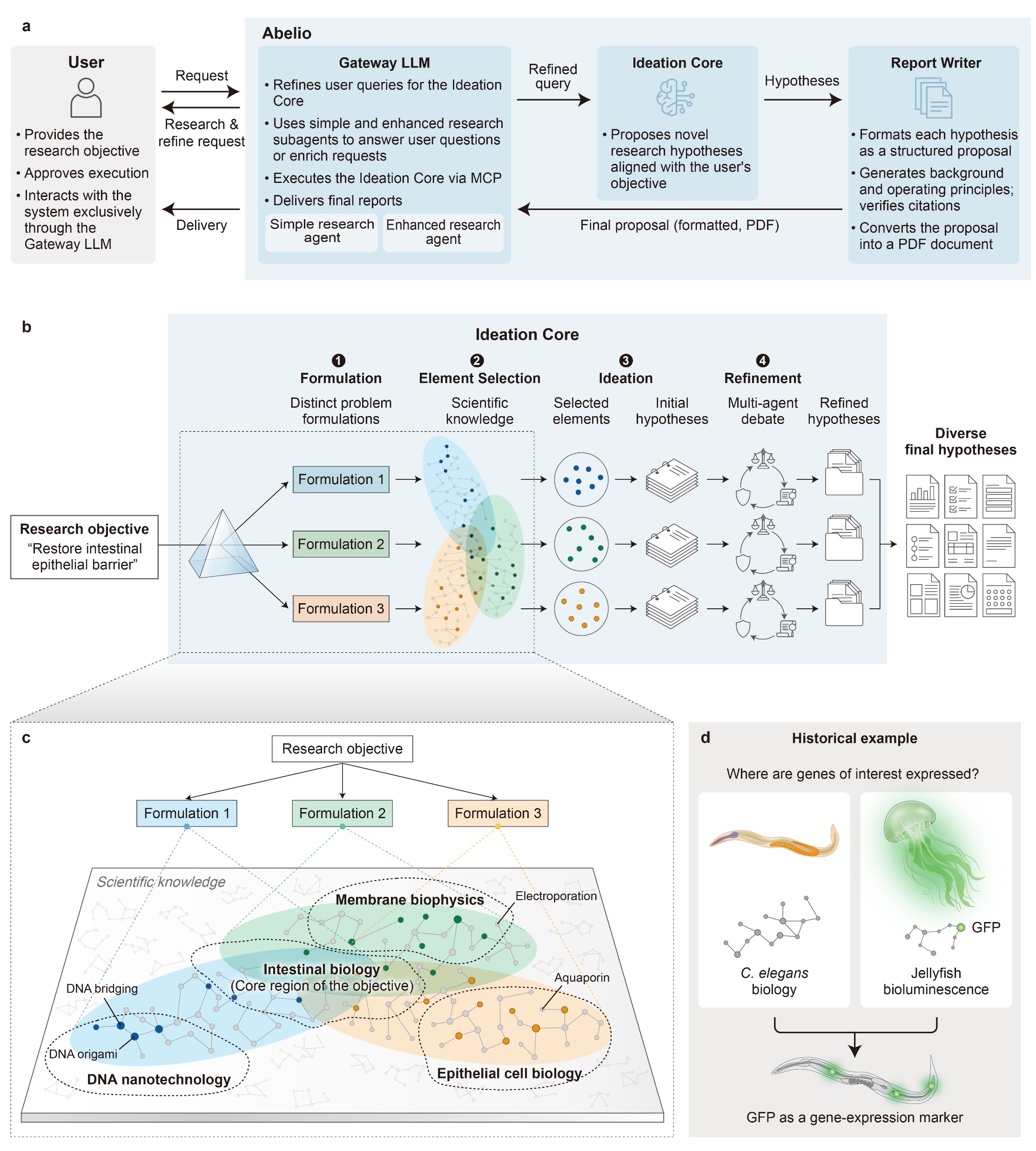}
  \caption[Abelio architecture]{%
    \textbf{Abelio architecture.}
    \textbf{a}, System overview. The Gateway Large Language Model (LLM) refines
    the user's research objective, invokes the Ideation Core through Model
    Context Protocol, and delivers the final PDF proposal generated by the
    Report Writer. Users interact exclusively through the Gateway LLM.
    \textbf{b}, Ideation Core workflow. Formulation reframes the objective from
    diverse perspectives. For each formulation, Element Selection identifies
    relevant elements from the literature, filters them for quality and novelty,
    and constructs an element pool. Ideation combines selected elements into
    initial hypotheses, which Refinement improves or discards through iterative
    multi-agent debate, producing up to ten diverse final hypotheses.
    \textbf{c}, Conceptual illustration of formulation-guided scientific
    knowledge exploration. Different formulations define overlapping search
    scopes spanning multiple research fields, revealing elements beyond those
    conventionally associated with the objective. Dashed boundaries represent
    research fields, shaded regions indicate formulation-specific scopes, and
    colored nodes indicate selected elements; colors distinguish formulations
    rather than fields.
    \textbf{d}, Historical example of cross-disciplinary knowledge transfer:
    application of jellyfish-derived green fluorescent protein as a
    gene-expression marker in \textit{C.\ elegans}.%
  }
  \label{fig:architecture}
\end{figure}

The system comprises three components (\cref{fig:architecture}a). The Gateway
LLM interacts with the researcher and translates a request into a refined
research objective, supported by two subagents conducting preliminary and
comprehensive literature surveys. Once approved by the researcher, the objective
enters the Ideation Core, which executes divergence--convergence cycles to
generate a set of hypotheses. The Report Writer subsequently converts each
retained hypothesis into a structured, referenced proposal, delivered as a PDF
for researcher evaluation. Detailed configurations are described in
\nameref{sec:methods}.

The Ideation Core consists of three subsystems (\cref{fig:architecture}b).
Formulation translates the research objective into multiple scientific
perspectives, generating distinct problem formulations before proposing
solutions. It then selects formulations spanning different regions of the
conceptual space (\cref{fig:architecture}c). Element Selection identifies
potential components capable of addressing each formulation. These comprise two
categories: material elements, including molecules, substances, and pathways and
strategy elements, representing overarching approaches to overcoming
bottlenecks or leveraging established mechanisms. Ideation \& Refinement selects
applicable elements from the resulting element pool and integrates them into
hypotheses extending beyond the objective's conventional solution space. The
historical example of GFP (\cref{fig:architecture}d) illustrates this process,
whereby elements originating from otherwise unrelated scientific contexts are
identified and repurposed to address a research objective. Each hypothesis
subsequently undergoes iterative debate among defender, reviewer, and judge
agents, resulting in refinement against an extensive literature corpus or
rejection. This process strengthens scientific soundness while sharpening the
distinguishing feature of each hypothesis.

Convergent selection during the expansion stages of Formulation and Element
Selection is performed by the Latent Selector. This component jointly reranks
candidates according to query alignment and inter-candidate diversity in latent
space, without invoking a generative model.

% ============== Results: System Evaluation and Analysis Subsection ============== %

\subsection{System evaluation and analysis}
\label{subsec:evaluation}

\textbf{Evaluation setup and cost.} We quantitatively assessed the research
hypotheses generated by Abelio using 50 research objectives curated by human
experts. Across these objectives, the Ideation Core generated 293 hypotheses, of
which 212 were retained by the Report Writer in the final reports. Nearly all
exclusions resulted from safety refusals triggered by biological content in the
underlying LLMs during the Formulation and report-writing stages. The complete
Abelio evaluation incurred a total cost of \$417.09, corresponding to an average
of \$1.97 per final proposal, including the costs associated with excluded
hypotheses. The Ideation Core incurred a cost of \$188.17 for 293 hypotheses
(averaging \$0.64 per hypothesis), whereas the Report Writer incurred a cost of
\$228.92 for 212 final proposals (averaging \$1.08 per proposal).

\textbf{Metrics.} Preliminary experiments revealed that rankings generated by
LLM judges were sensitive to variations in prompts, judge models, and scoring
schemes. Previous studies have similarly raised concerns regarding the
reliability of LLM-based evaluations of scientific ideas, particularly in
assessing
novelty\autocite{baumann2026stopautomatingpeerreview,sinhahajari2026limitsllmasjudgescientificnovelty}.
To address these limitations, we established three evaluation metrics:
inter-hypothesis diversity, literature distance, and scientific soundness.
Inter-hypothesis diversity quantifies differences among hypotheses generated for
a given research objective, whereas literature distance measures the divergence
of a proposed hypothesis from the literature retrieved for that objective.
Scientific soundness is assessed using an evaluation framework adapted from
LiveIdeaBench\autocite{ruan2026evaluating}, which scores groundedness,
sharpness, and feasibility. The first two metrics are computed without
judge-based assessment to minimize potential human and LLM biases, whereas
scientific soundness requires contextual evaluation and is therefore assessed
using an LLM judge. Detailed implementations of these metrics are provided in
\nameref{sec:methods}.

\textbf{Comparison with frontier LLMs.} We initially benchmarked the Ideation
Core against frontier LLMs using the same 50 research objectives to assess the
quality of its generated hypotheses. Because the Ideation Core produces
substantially longer outputs than frontier LLMs, we summarized its hypotheses to
standardize length, structure, and writing style, thereby minimizing their
influence on the evaluation (\snoteone). Four frontier models were
assessed using identical objectives: GPT-5.2-high, Gemini 3.1 Pro Preview,
Claude Opus 4.6, and Grok 4.1 Fast Reasoning. To maintain consistency with
Abelio's model cutoff, we used the versions available in February 2026, with
built-in web search enabled for each model (\nameref{sec:methods}).

The Ideation Core achieved significantly higher mean scientific soundness scores
than all four models and greater mean literature distance than three models
(\cref{fig:evaluation}a). Gemini 3.1 Pro was the sole exception, showing no
significant difference in mean literature distance from the Ideation Core.
Literature distance and scientific soundness represent distinct dimensions of
research hypothesis quality. However, hypotheses may deviate substantially from
published research because they lack scientific rigor, whereas well-grounded
hypotheses may offer only incremental advances. \Cref{fig:evaluation}a
illustrates both dimensions simultaneously, showing that hypotheses generated by
the Ideation Core occupy the upper-right region of the plot.
\Cref{fig:evaluation}b presents the proportion of hypotheses exceeding varying
thresholds for the combined standardized literature distance and scientific
soundness scores. The Ideation Core consistently outperformed all four models
across these thresholds. Furthermore, it generated more diverse sets of
hypotheses for each objective than any frontier LLM (\cref{fig:evaluation}c).

\subsection[Each step contributes to its targeted property]{Each step in the
  Ideation Core contributes to its targeted property}
\label{subsec:ablation}

\begin{figure}[p]
  \centering
  \includegraphics[width=\linewidth]{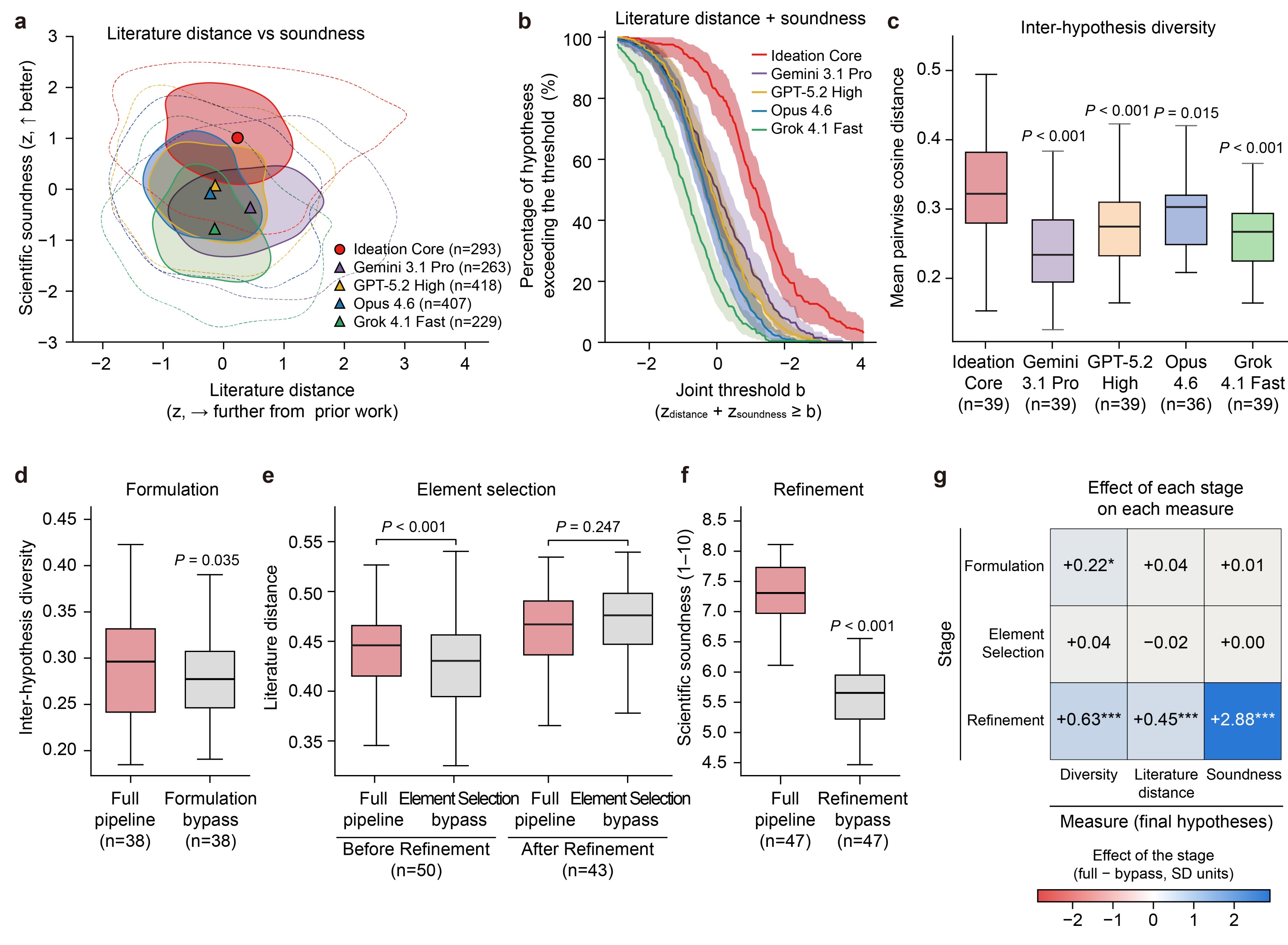}
  \caption[Ideation Core comparison with frontier language models and stage
  ablations]{%
    \textbf{Ideation Core comparison with frontier language models and stage
    ablations.}
    \textbf{a}, Literature distance versus scientific soundness for hypotheses
    generated by the Ideation Core ($n = 293$), Gemini 3.1 Pro ($n = 263$),
    GPT-5.2 High ($n = 418$), Opus 4.6 ($n = 407$), and Grok 4.1 Fast ($n =
    229$) across the same 50 research objectives. Literature distance represents
    the mean embedding distance between a hypothesis and the ten papers
    retrieved for its objective; soundness is the mean score from three LLM
    judges on a ten-point scale. Both axes are z-scored across systems. Filled
    and dashed contours enclose the densest 50\% and 90\% of hypotheses per
    system, respectively; markers indicate means.
    \textbf{b}, Proportion of hypotheses with summed z-scores from \textbf{a}
    exceeding threshold $b$ as $b$ increases; bands indicate 95\% bootstrap
    intervals from objective resampling.
    \textbf{c}, Inter-hypothesis diversity, calculated as the mean pairwise
    embedding distance among hypotheses per objective.
    \textbf{d--f}, Single-stage ablations comparing the full pipeline with
    individual stage bypasses: \textbf{d}, Formulation (diversity);
    \textbf{e}, Element Selection (literature distance before [left] and after
    [right] Refinement); and \textbf{f}, Refinement (soundness).
    \textbf{g}, Stage effects across all measures, expressed as median
    within-objective differences (full pipeline minus bypass) standardized by
    the full pipeline's across-objective standard deviation; outlined cells
    correspond to tests in \textbf{d}, \textbf{e} (right) and \textbf{f}.
    Panels \textbf{a--c} use Ideation Core hypothesis summaries matched to
    frontier-model outputs in length and style; \textbf{d--f} use unmodified
    hypotheses. Boxes indicate medians and interquartile ranges (IQRs) across
    objectives; whiskers extend to the most extreme values within $1.5 \times$
    IQR. $p$ values were calculated using two-sided paired Wilcoxon signed-rank
    tests on per-objective means, with Holm adjustment across four frontier
    models in \textbf{c}, three stages in \textbf{d--f}, and six remaining
    comparisons in \textbf{g}; \textbf{e} (left) was unadjusted. In \textbf{g},
    $^{*}p < 0.05$, $^{***}p < 0.001$. Values of $n$ beneath boxes indicate
    objectives covered by both compared systems; diversity requires at least two
    hypotheses per objective.%
  }
  \label{fig:evaluation}
\end{figure}

We conducted a pipeline-level ablation analysis by subsequently removing the
Formulation, Element Selection, and Refinement steps. Each variant was compared
with the complete pipeline using the metric corresponding to the intended
contribution of the respective step: inter-hypothesis diversity for Formulation,
literature distance for Element Selection, and scientific soundness for
Refinement (\cref{fig:evaluation}d--f). To distinguish the contribution of each
step from run-to-run variability, we reused cached upstream outputs and
recomputed only the ablated step and subsequent components. Removing Formulation
significantly reduced inter-hypothesis diversity (\cref{fig:evaluation}d),
indicating its contribution to expanding the range of approaches generated per
objective. Omitting Element Selection did not significantly alter literature
distance relative to the complete pipeline (\cref{fig:evaluation}e, right). We
therefore compared the complete pipeline with the Element Selection bypass
before Refinement and found that bypassing Element Selection significantly
reduced literature distance (\cref{fig:evaluation}e, left). This finding
indicates that Element Selection directs hypotheses beyond conventional
approaches, although Refinement largely obscures this effect, leaving no
statistically significant difference in the final literature distance metric.
Removing Refinement significantly reduced scientific soundness
(\cref{fig:evaluation}f), demonstrating its contribution to the scientific rigor
of the final proposals. \Cref{fig:evaluation}g summarizes all stage-metric
comparisons. No ablation variant significantly outperformed the complete
pipeline on metrics outside the targeted property of the corresponding step.
Collectively, these findings support the contributions of Formulation to
hypothesis diversity and Refinement to scientific soundness, whereas the
influence of Element Selection on literature distance was detectable before, but
not after, Refinement.

% ============== Results: From Objective to Testable Hypotheses Subsection ============== %

\subsection{From objective to testable hypotheses}
\label{subsec:testable}

To evaluate whether the generated proposals could withstand experimental
testing, we conducted wet-lab validation of representative proposals. Given the
objective of restoring the intestinal epithelial barrier, particularly tight
junction integrity, Abelio generated nine candidate proposals (see
\nameref{sec:data-availability}). One proposal diverged substantially from
conventional approaches. Rather than biochemically targeting the regulation,
expression, or assembly of tight junction
proteins\autocite{graham2019intracellular,schuppan2021randomized,ghosh2021regulation},
it proposed physically narrowing the intercellular junctional gap. The proposal
contrasted luminal application of free poly(ethylene glycol) (PEG), intended to
narrow the cleft through polymer crowding, with membrane-anchored complementary
DNA strands designed to form intercellular nanobridges. Although PEG-mediated
effects on permeability have been reported, their independence from signaling
mechanisms remains incompletely
established\autocite{wu2004high,valuckaite2009oral,bharadwaj2011higher,casselbrant2018luminal,gao2023dspe},
and we identified no precedent for DNA nanobridges in barrier restoration.
Neither strategy had previously been considered by the investigators who defined
the objective.

\begin{figure}[p]
  \centering
  \includegraphics[width=\linewidth]{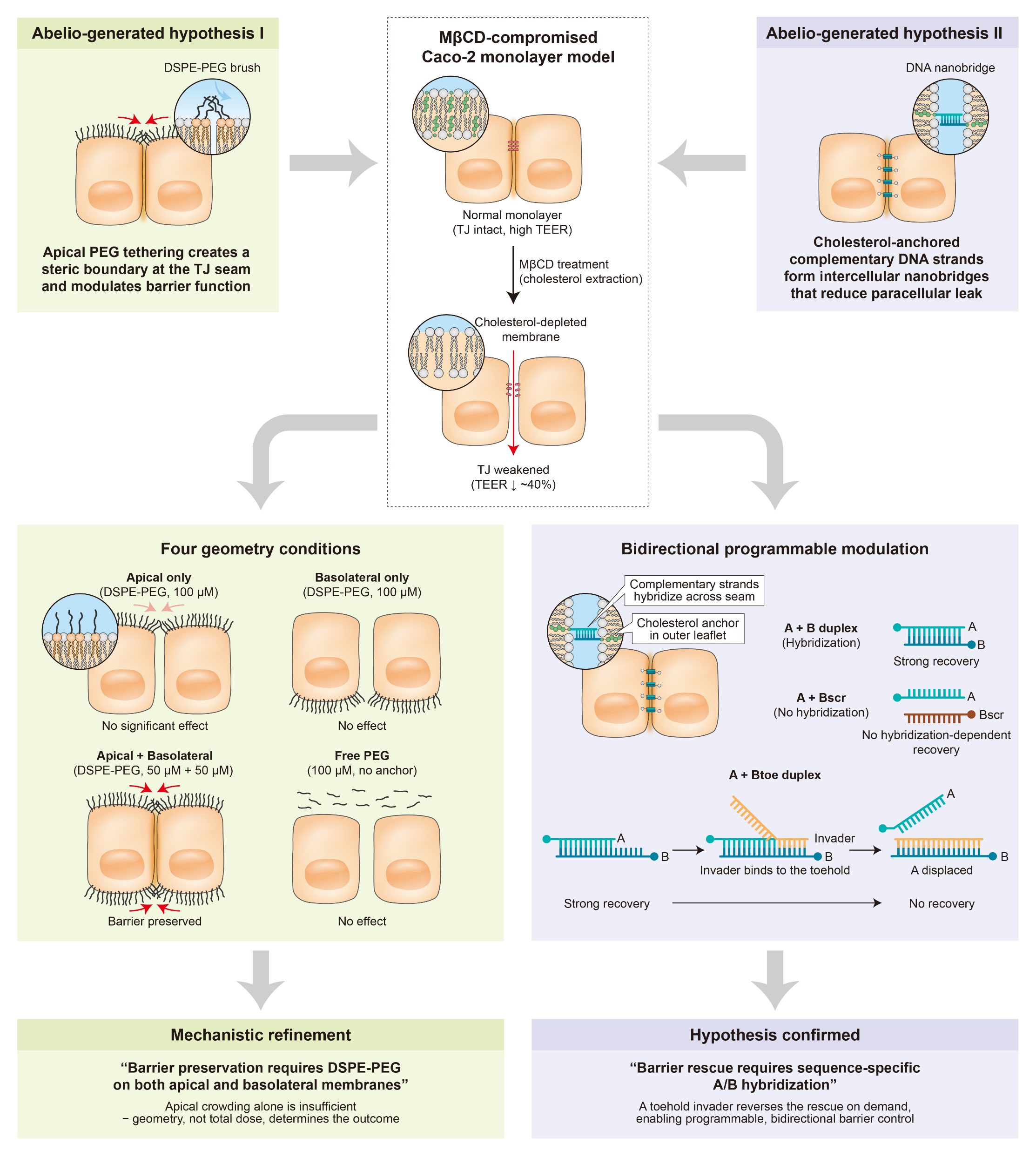}
  \caption[From Abelio-generated hypotheses to experimental outcomes]{%
    \textbf{From Abelio-generated hypotheses to experimental outcomes.}
    Top: Two Abelio-generated hypotheses for intestinal barrier restoration,
    flanking the shared experimental model. Left, apical poly(ethylene glycol)
    (PEG) tethering to create a steric boundary at the tight junction seam.
    Center, methyl-$\beta$-cyclodextrin (M$\beta$CD)-mediated cholesterol
    depletion in mature Caco-2 monolayers, reducing transepithelial electrical
    resistance (TEER) by approximately 40\%. Right, intercellular nanobridges
    formed by cholesterol-anchored complementary DNA. Middle: experimental
    conditions for each hypothesis, showing four PEG tethering geometries (left)
    and hybridization and toehold-displacement conditions (right). Bottom:
    Conclusions for each strategy. PEG brushes and DNA strands spanning the
    paracellular seam depict proposed configurations rather than directly
    observed structures.%
  }
  \label{fig:proposals}
\end{figure}

The investigators subsequently translated the selected report into two follow-up
objectives, focusing separately on PEG and complementary DNA
(\snotetwo). Each objective yielded eight reports (see
\nameref{sec:data-availability}), from which the investigators selected one
proposal for experimental evaluation. The PEG proposal specified comparisons
among apical, basolateral, and symmetric (both surfaces) tethering, using free
PEG as a non-tethered control. The DNA nanobridge proposal preserved
complementary cholesterol-modified DNA strands as its central principle while
defining experimental conditions and controls, including scrambled
oligonucleotides and cholesterol repletion (\cref{fig:proposals}; complete
proposal in \sappone).

Both strategies were then prospectively evaluated \textit{in vitro} using Caco-2
monolayers cultured on Transwell inserts for 21 days. Abelio specified a target
barrier impairment corresponding to an approximately 40\% reduction in
transepithelial electrical resistance (TEER). We empirically established the
conditions required to achieve this reduction, selecting treatment with 20 mM
methyl-$\beta$-cyclodextrin (M$\beta$CD) for 1 h based on a dose--response
analysis (\sfigref{sfig:mbcd}a). This treatment reduced TEER by approximately 40\%
without compromising cell viability (\sfigref{sfig:mbcd}b), establishing a
partially impaired, submaximal barrier state suitable for detecting both further
deterioration and recovery.

% ============== Results: PEG Strategy Subsection ============== %

\subsection[PEG treatment of both surfaces improves barrier readouts]{PEG
  treatment of both surfaces improves barrier readouts in damaged Caco-2
  monolayers}
\label{subsec:peg}

Abelio predicted that apical PEG tethering alone would establish sufficient
steric boundaries at the tight junction interface, based on the localization of
tight junctions to the apical-most region of the lateral membrane. Abelio
identified 1,2-distearoyl-\textit{sn}-glycero-3-phosphoethanolamine
(DSPE)-PEG as the candidate polymer, and we empirically determined the working
concentrations. Concentrations were selected based on viability assays for
DSPE-PEG (\sfigref{sfig:controls}a) and free PEG (\sfigref{sfig:controls}b), which
revealed no cytotoxicity across the tested ranges. Initial comparisons of three
tethering configurations at 50 $\mu$M per surface demonstrated that neither
apical-only (Api, 50 $\mu$M) nor basolateral-only (Baso, 50 $\mu$M) treatment
preserved TEER. In contrast, combined apical and basolateral tethering
maintained significantly higher TEER than M$\beta$CD treatment alone at
concentrations starting from 25 $\mu$M per side (\sfigref{sfig:controls}c). We
therefore increased the single-surface doses to 100 $\mu$M to match the total
exposure of the dual-surface condition (50 $\mu$M each side).

Contrary to the specific prediction, the observed effect did not depend on
apical crowding alone. Dual-surface tethering (50 $\mu$M per side) maintained
significantly higher TEER than M$\beta$CD treatment alone, under which TEER
continued to decline (\cref{fig:wetlab}a). Although apical-only treatment
exhibited a trend toward higher TEER than basolateral-only treatment, neither
significantly improved TEER relative to M$\beta$CD treatment alone in the
dose-finding or dose-matched experiments (\cref{fig:wetlab}a,
\sfigref{sfig:controls}c). These findings suggest that Abelio identified a
promising intervention but proposed an incomplete mechanism. Free PEG had no
effect at 100 $\mu$M, matching the total exposure of the dual-surface condition
(\cref{fig:wetlab}b), with a modest effect observed only at 500 $\mu$M, beyond
the matched comparison. In the paracellular permeability assay, conducted using
the 2 h dextran exposure established in pilot experiments
(\sfigref{sfig:controls}d), dual-surface tethering reduced fluorescein
isothiocyanate (FITC)--dextran flux relative to M$\beta$CD treatment alone,
whereas free PEG at the matched concentration had no effect
(\cref{fig:wetlab}c). Thus, at matched exposure, membrane anchoring, rather than
PEG alone, was critical for preserving barrier function.

Fluorescence intensity profiles revealed rhodamine signal throughout the
apical-to-basal extent following dual-surface treatment, whereas apical-only and
basolateral-only treatments produced weaker signals across the intervening
lateral region (\cref{fig:wetlab}d). Neither untreated nor M$\beta$CD-only
monolayers exhibited detectable rhodamine fluorescence
(\sfigref{sfig:controls}e), confirming that the signal in \cref{fig:wetlab}d
originated from the administered PEG-lipid rather than cellular
autofluorescence. Whether widened paracellular clefts at damaged sites permit
luminally administered PEG-lipids to access both apical and basolateral
surfaces, potentially facilitating preferential delivery to regions of barrier
disruption, remains to be investigated.

% ============== Results: DNA Nanobridge Subsection ============== %

\subsection[DNA nanobridges produce sequence-dependent, reversible effects]{DNA
  nanobridges produce sequence-dependent and reversible barrier effects}
\label{subsec:dna}

Abelio proposed that cholesterol-anchored complementary DNA strands could form
intercellular nanobridges to physically restrict paracellular leakage, an
approach without precedent in tight junction biology. Following Abelio's
specifications (7--9 bp duplexes, binding free energy of $-10$ to
$-15$ kcal/mol to
enable transient but repeatedly reforming bridges), we designed complementary
8-nt cholesterol--tetraethylene glycol (Chol-TEG)-modified oligonucleotides
(Chol-TEG-A and Chol-TEG-B, hereafter A and B) which were sequentially applied
to apical membranes to form intercellular bridges (see \nameref{sec:methods} for
sequences). The resulting 8-bp duplex had a predicted binding free energy of
$-11.9$ kcal/mol, consistent with Abelio's specified range. Each strand was
administered at 0.5 $\mu$M, maintaining total oligonucleotide exposure at $\le$
1 $\mu$M, a concentration range that did not affect monolayer viability
(\sfigref{sfig:controls}f). To assess hybridization specificity, Abelio specified a
scrambled B strand control (Bscr) with identical base composition but a
non-complementary sequence.

\begin{figure}[p]
  \centering
  \includegraphics[width=\linewidth]{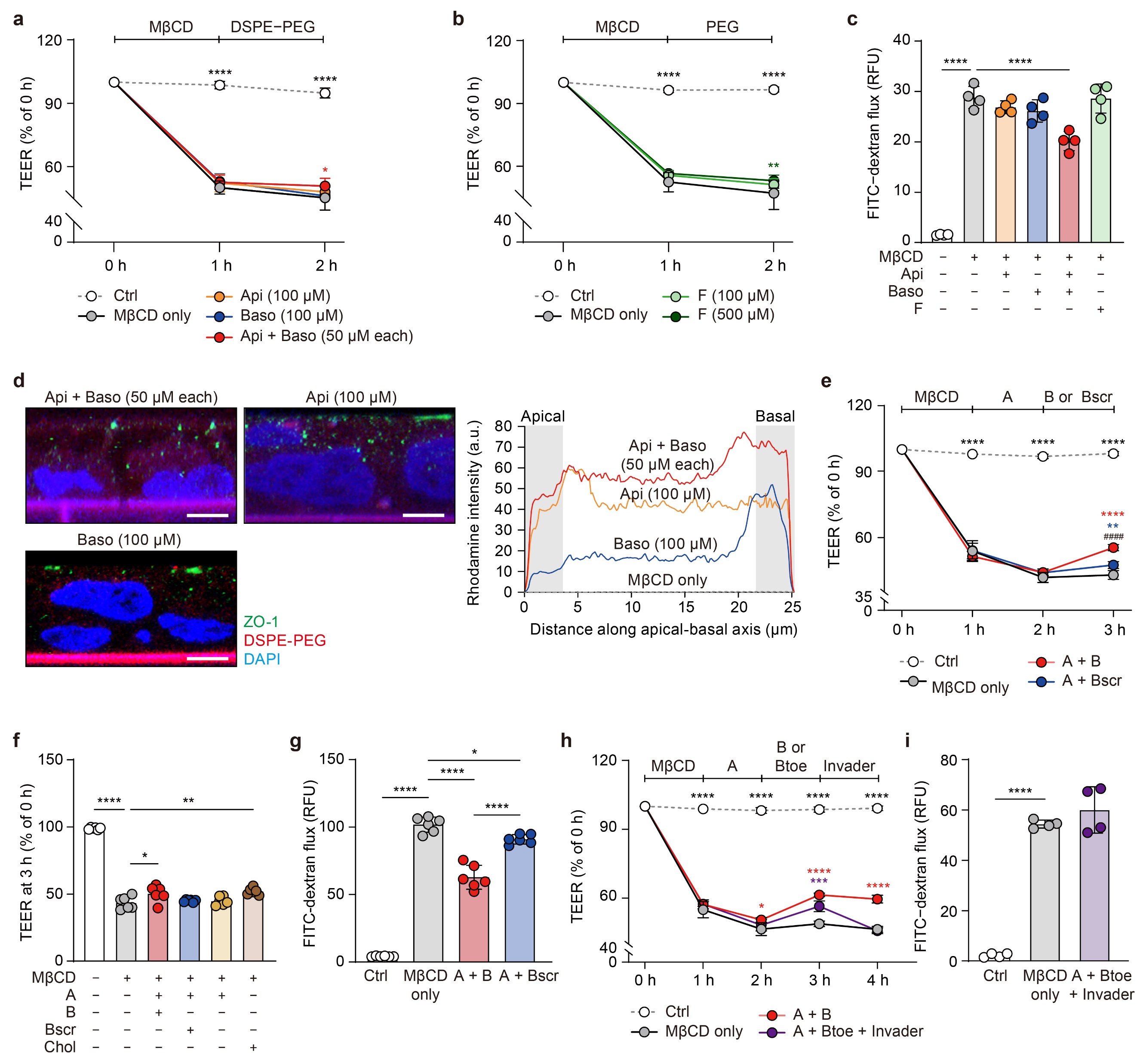}
  \caption[Experimental validation]{%
    \textbf{Experimental validation.}
    \textbf{a}, TEER kinetics of M$\beta$CD-compromised Caco-2 monolayers
    following dose-matched
    1,2-distearoyl-\textit{sn}-glycero-3-phosphoethanolamine-poly(ethylene
    glycol) (DSPE-PEG) tethering: apical (Api, 100 $\mu$M), basolateral (Baso,
    100 $\mu$M), or both surfaces (Api + Baso, 50 $\mu$M each).
    \textbf{b}, TEER kinetics following non-tethered free PEG treatment
    (F 100, F 500; 100 and 500 $\mu$M).
    \textbf{c}, Fluorescein isothiocyanate (FITC)--dextran paracellular flux
    across tethering conditions, measured as basolateral fluorescence 2 h after
    apical application (excitation/emission: 485/535 nm). RFU, relative
    fluorescence units.
    \textbf{d}, Confocal Z-stack XZ reconstructions showing rhodamine-DSPE-PEG
    distribution and apical-to-basal fluorescence intensity profiles. Shading
    indicates monolayer boundaries. Scale bars, 5 $\mu$m.
    \textbf{e}, TEER kinetics following sequential application of
    cholesterol-tetraethylene glycol-modified DNA strand A and complementary B
    or scrambled B (Bscr) after 1 h. $^{\#\#\#\#}p < 0.0001$, A + B versus
    A + Bscr at 3 h.
    \textbf{f}, TEER at 3 h (\% of 0 h), comparing nanobridge conditions, A
    alone, and cholesterol repletion (Chol, cholesterol-loaded M$\beta$CD).
    \textbf{g}, FITC--dextran flux across nanobridge conditions.
    \textbf{h}, TEER kinetics following A, toehold-bearing B (Btoe), and
    unmodified invader (A + Btoe + Invader), compared with toehold-free A + B.
    \textbf{i}, FITC--dextran flux for control (Ctrl), M$\beta$CD-only, and
    A + Btoe + Invader.
    Unless indicated, comparisons are against M$\beta$CD-only, marked by
    color-matched asterisks (\textbf{a,b,e,h}) or brackets
    (\textbf{c,f,g,i}). $n = 6$ wells/condition from two independent experiments
    (\textbf{a, b, e, f, g, h}); $n = 4$ wells/condition from two independent
    experiments (\textbf{c, i}). In \textbf{d}, three fields/well, two
    wells/condition, and two independent experiments were analyzed.
    $^{*}p < 0.05$, $^{**}p < 0.01$, $^{***}p < 0.001$, and
    $^{****}p < 0.0001$; two-way repeated-measures ANOVA with Dunnett's
    (\textbf{a, b, h}) or Tukey's (\textbf{e}) tests; one-way ANOVA with
    Dunnett's (\textbf{c, f, i}) or Bonferroni's (\textbf{g}, prespecified
    pairs) tests.%
  }
  \label{fig:wetlab}
\end{figure}

The A + B condition significantly increased TEER relative to M$\beta$CD
treatment alone, with recovery significantly exceeding that observed with
A + Bscr (\cref{fig:wetlab}e), demonstrating that effective barrier restoration
depended on sequence-specific hybridization. A + Bscr produced a smaller
increase relative to M$\beta$CD treatment alone, consistent with a potential
contribution of cholesterol anchors to membrane lipid content independently of
duplex formation. At 3 h, TEER recovery with A + B was comparable to that
achieved by cholesterol repletion, a benchmark for restoring native lipid-raft
architecture, whereas neither A alone nor A + Bscr approached this level
(\cref{fig:wetlab}f). FITC--dextran flux further supported sequence dependence,
with lower permeability under A + B than A + Bscr (\cref{fig:wetlab}g).

To further validate reversibility, we tested a modified B strand containing a
6-nt toehold overhang (Btoe, 14 nt)\autocite{zhang2009control}. A + Btoe
achieved TEER recovery comparable to A + B, whereas subsequent addition of an
unmodified 14-nt invader strand reduced TEER toward the M$\beta$CD-only level.
In contrast, the toehold-free A + B condition, which resists strand
displacement, maintained its recovery (\cref{fig:wetlab}h). Consistently,
FITC--dextran flux following invader addition was indistinguishable from that
observed with M$\beta$CD treatment alone (\cref{fig:wetlab}i), further
supporting reversal of the barrier-protective effect.

As with PEG-lipids, widened paracellular clefts in compromised monolayers may
permit apically administered oligonucleotides to access lateral membranes,
although duplex formation was not directly visualized. Because hybridization
confined to a single membrane would not reduce paracellular spacing, the
sequence-specific and reversible changes in TEER and dextran permeability are
most consistent with duplex formation between opposing membranes. Collectively,
these findings demonstrate sequence-specific, externally reversible modulation
of epithelial barrier function using cholesterol-anchored complementary DNA.

% ============== Discussion Section ============== %

\section{Discussion}
\label{sec:discussion}

Abelio was developed to expand the range of research directions considered for a
given objective. Its architecture explicitly incorporates the exploration of
alternative scientific perspectives into hypothesis generation, using distinct
problem formulations to guide the identification of relevant elements and
development of candidate approaches. Divergent expansion and diversity-aware
selection preserve multiple possibilities for subsequent refinement. The
resulting proposals are designed to support investigators in identifying and
developing research directions, with their value evaluated through both the
diversity of approaches generated and the experiments they enable.

The quantitative findings support this intended function. The greater diversity
of Abelio-generated proposals, together with the reduction observed upon
removing Formulation, demonstrates the value of exploring alternative
perspectives before hypothesis construction. Higher scientific soundness scores
further suggest that this diversity was accompanied by proposals evaluated as
scientifically grounded and feasible, while the Refinement ablation supports the
contribution of subsequent scrutiny to proposal quality. Removing Element
Selection reduced literature distance before Refinement but not in the final
proposals, suggesting that this step directs hypotheses beyond established
approaches during their initial construction. As Refinement modifies each
hypothesis and independently retrieves literature, it may diminish the
difference introduced by Element Selection. Moreover, bypassing Element
Selection also modifies the ideation prompt; thus, the pre-Refinement difference
reflects the combined effects of both changes.

The intestinal barrier case illustrates the progression of an Abelio-generated
direction from investigator selection through two follow-up proposals to
experimental validation. These experiments demonstrate how expanding the range
of approaches can create additional avenues for investigation. By targeting
barrier restoration through physical interactions at cell surfaces, the
proposals extended the intervention strategies beyond conventional approaches.
The PEG and DNA strategies yielded informative results in distinct ways.
Although the prediction that apical PEG tethering alone would restore barrier
function was not supported, the proposed comparison of tethering configurations
identified dual-surface tethering as an important condition for preventing
further barrier deterioration. For the DNA nanobridge strategy,
complementarity-dependent recovery and its reversal following invader addition
supported the proposed application of membrane-anchored DNA strands to regulate
barrier function, demonstrating programmable and reversible modulation of
epithelial permeability. Collectively, these findings show that proposed
research directions can be valuable both when experiments validate their central
principles and when they identify conditions requiring revision of the initial
design. Mechanistically, both interventions exerted stronger effects on
macromolecular flux than on transepithelial resistance. TEER primarily reflects
the claudin-mediated pore pathway, whereas 4 kDa dextran traverses the larger
leak pathway\autocite{horowitz2023paracellular}. Physical narrowing of the
paracellular space would preferentially restrict the latter, consistent with our
observations, although attributing these effects to a specific pathway requires
measurements using solutes of varying sizes.

DNA nanotechnology is an active research field with applications in modulating
cell-surface and cell--cell
interactions\autocite{seeman2017dna,pinheiro2011challenges,schoenit2021functionalization}.
While previous studies have investigated applications involving simple reporter
probes, suspension cells, and immune
interactions\autocite{gartner2009programmed,schoenit2021tuning,kim2026amphiphilic,you2017dna,ge2020programming,hoffecker2019tuning},
our experiments evaluated barrier responses in damaged intestinal epithelial
monolayers. Extending these principles to barrier restoration demonstrates how
knowledge originating in a distinct research context can address an existing
scientific objective, illustrating Abelio's potential to expand the approaches
available to investigators.

These experimental findings provide proof of concept for developing
Abelio-generated research directions into testable interventions. Further
investigations, including \textit{in vivo} validation, are necessary to
establish the practical utility of these interventions, while evaluation across
additional research objectives is required to determine the generalizability of
the approach. Translating these concepts into experiments also required domain
expertise. Investigators selected and refined the PEG and DNA strategies, while
Abelio provided strategic designs and experimental recommendations. Experimental
implementation required investigator-led decisions and empirical optimization,
including oligonucleotide sequence selection, working concentrations, and assay
conditions. Thus, the resulting experiments represent a collaborative process in
which AI-generated directions were refined and evaluated through continued
scientific judgment.

More broadly, reliable assessment of scientific hypothesis novelty and quality
remains challenging. Formal verification can establish whether mathematical
conclusions follow from specified assumptions. In contrast, empirical hypotheses
concern the natural world and require experimental evidence beyond logical
consistency or agreement with existing literature to establish their validity.
Obtaining conclusive evidence may also be limited by available methodologies and
instrumentation. Consequently, our metrics assess specific dimensions of
proposal quality rather than providing comprehensive validation. Furthermore,
LLM-based judges have recognized
limitations\autocite{baumann2026stopautomatingpeerreview,sinhahajari2026limitsllmasjudgescientificnovelty,jiang2025artificial},
and no individual metric can determine whether a proposal will yield a valuable
experiment or significant discovery. Quantitative evaluations must therefore be
considered alongside expert assessment and, where feasible, prospective
experimental validation. Developing more reliable methods to evaluate the
novelty and quality of scientific hypotheses remains an unresolved challenge
essential to advancing AI-driven scientific research.

Overall, Abelio offers a structured framework for expanding the scientific
possibilities considered for a research objective. Its quantitative evaluation
and prospective experiments support the value of generating diverse approaches
that investigators can evaluate, refine, and develop. Such systems can advance
scientific discovery by expanding the research directions available for human
evaluation and facilitating their translation into testable hypotheses.

% ==========================================================
% Methods
% ==========================================================

% ============== Methods Section ============== %

\section{Methods}
\label{sec:methods}

\subsection{Abelio architecture}

Each agent in Abelio uses a commercial model from OpenAI, Anthropic, Google, or
xAI, with role-optimized prompts and literature-retrieval tools or
provider-native web search as required. Researchers submit requests through a
chat interface to the Gateway LLM, which refines them into research objectives
for the Ideation Core. For each objective, the system generates multiple
hypotheses, which the Report Writer compiles into a formatted report. The
Gateway LLM then returns the report with summaries of each idea.

\subsection{Gateway LLM}

The Gateway LLM employs a simple research agent for preliminary literature
searches and an enhanced research agent for in-depth, iterative investigation of
specific research topics. Through this process, the Gateway LLM helps
researchers
assess the current state of the field and refine research objectives. Each
objective comprises a concise core goal defining the intended outcome and key
constraints specifying acceptable solutions, including technical requirements,
methodological preferences, and restrictions.

\subsection{Ideation Core}

The Ideation Core generates and refines hypotheses from the research objective
through three subsystems: Formulation, Element Selection, and Ideation \&
Refinement.

\textbf{Formulation}. Three agents operate concurrently from distinct
perspectives: biological scales ranging from molecular to systemic, alternative
mechanistic pathways and associated biological entities, and unresolved
questions, inconsistencies, and documented limitations in the literature.
Independent agents then integrate these perspectives, incorporate strategic
viewpoints, and consolidate them into distinct problem formulations. The Latent
Selector reranks these formulations, selecting 10 that align with the research
objective while remaining mutually distinct.

\textbf{Element Selection}. Major and minor requirements are extracted from each
formulation to define the search space for strategy and material elements.
Agents first generate strategy elements inspired by approaches from other
domains addressing similar problems. Implementation-specific details are removed
to retain core principles, followed by cross-domain inspiration to expand the
search space. Agents subsequently identify materials satisfying the requirements
through literature and web searches. Independent agents identify alternatives
for each material, which the Latent Selector reranks to minimize redundancy and
maximize search-space coverage. Material and strategy elements are subsequently
filtered for novelty, objective-specific constraints, and formulation
requirements to ensure novelty and applicability. LLM agents perform this
context-dependent filtering. To avoid bias toward conventional solutions, search
agents receive only requirements, not the formulation or original objective.
Finally, material and strategy elements are combined into an element pool for
each formulation.

\textbf{Ideation \& Refinement}. Each ideation instance randomly samples 75\% of
the element pool to mitigate the LLM hivemind effect\autocite{jiang2025artificial},
whereby models favor familiar elements over potentially promising alternatives.
We observed strong preferences for concepts such as liquid--liquid phase
separation (LLPS) and microfluidic chips among recent frontier LLMs. The
ideation agent selects two to three elements from the subset and generates a
hypothesis satisfying the objective within the formulation. The proposal then
undergoes debate among defender, reviewer, and judge agents to improve its
scientific soundness and novelty or reject scientifically unsound hypotheses.
The reviewer critiques the proposal using literature searches, and the defender
responds with literature-supported arguments and a refined proposal or discards
it if refinement is infeasible. After two warm-up rounds, the judge determines
whether to continue, reject, or approve the proposal. Debate continues for up to
seven rounds; in the final round, the defender considers the last critique and
either discards the proposal or submits it to the Report Writer. The defender
retains memory throughout, whereas reviewers and judges remain stateless to
ensure independent evaluation of each revision.

\subsection{Report Writer}

The report writer processes each surviving hypothesis. A writing agent generates
a title, a concise concept summary (abstract), background contextualizing prior
work and unresolved limitations, and a proposed mechanism of action. A second
agent adds a decision framework specifying a binary success criterion,
mechanistic success and failure conditions, and criteria for pursuing, revising,
or abandoning the proposal. Citation markers are matched with retrieved
literature and compiled into a reference list, with unresolved markers flagged
and removed to ensure all retained citations correspond to existing
publications. Literature metadata are retrieved using
OpenAlex\autocite{priem2022openalex}. The system sends a typeset PDF link and
Markdown summary of each proposal to the Gateway LLM for delivery to the
researcher. Researchers can subsequently screen proposals using the summaries
and consult full reports to assess their experimental potential.

\subsection{Cross-cutting: caching and cost accounting}

Computationally intensive operations use a content-addressed cache keyed by
agent configuration fingerprints and input content. Changes to prompts, models,
or inputs invalidate only affected entries, enabling resumable execution and
attribution of output differences to specific configuration changes. Token usage
for each model call is recorded with its cache entry, enabling calculation of
total pipeline cost (summed across all entries) and incremental run cost
(counting only cache misses triggering model calls).

\subsection{Latent selector}

Each subsystem expands and contracts a candidate set, with retained candidates
defining the subsequent search space. Although LLM-based selection is possible,
positional and verbosity biases favor similar candidates, reducing diversity.
Abelio instead uses the Latent Selector, which numerically optimizes query
alignment and inter-candidate diversity through maximal marginal relevance. A
bi-encoder (Qwen3-Embedding-8B) embeds and L2-normalizes candidates, which
$k$-means clustering partitions into $k$ clusters, where $k$ equals the desired
number of retained candidates. Within each cluster, a cross-encoder
(zerank-2-reranker) scores candidates against the query, selecting the
highest-scoring representative. These representatives constitute the final
selection. Clustering preserves coverage of distinct candidate regions, while
cross-encoder ranking selects the strongest representative from each region,
avoiding biases associated with LLM-based selection.

\subsection{Metrics for system evaluation and analysis}

Inter-hypothesis diversity is the mean pairwise cosine distance among hypotheses
generated for an objective, reflecting the breadth of proposed approaches; it is
undefined when the number of hypotheses is below two. Literature distance is the
mean cosine distance between a hypothesis and the 10 published abstracts
retrieved for its objective, measuring semantic divergence from the literature.
Cosine distances were calculated using 4096-dimensional, L2-normalized
Qwen3-Embedding-8B embeddings, with hypotheses and abstracts encoded as
documents and objectives as instruction-conditioned queries. Abstracts were
retrieved from OpenAlex using hybrid BM25 and vector search, reranked with
zerank-2-reranker, and restricted to the top 10.

LLM judges assessed only scientific soundness, which requires interpreting
hypotheses within their objectives. Novelty was excluded because LLM judges may
reward novelty-like phrasing rather than genuine novelty. Soundness comprises
three criteria, each scored from 1 to 10: \textbf{grounding}, whether claims
rely on established mechanisms or valid reasoning rather than superficial
correlations or unverifiable premises; \textbf{sharpness}, whether all elements
support the central claim rather than being added arbitrarily to imply novelty,
without penalizing stated contingencies or potential experimental outcomes;
\textbf{feasibility}, whether the proposed design is achievable using current
methods, with penalties for overengineering. Judges assessed only these
dimensions, excluding overall quality and novelty, and received each hypothesis
and its objective without identifying the generating arm
(\snoteone). Soundness was calculated as the mean of the three
criteria, averaged across judges.

Following an adapted LiveIdeaBench protocol, three judges per hypothesis were
sampled from an LLM pool to limit dependence on individual
models\autocite{ruan2026evaluating}. Seeded shuffling ensured deterministic
assignments. Models included in the frontier comparison (described below) were
excluded from judging. We selected the strongest available models not included
in the evaluation, occasionally opting for lower-tier models or model families
rather than developers' current flagships. The pool comprised Claude Opus 4.8,
GPT-5.5-high, Gemini 3.5 Flash, Grok 4, DeepSeek 4 Pro, Qwen3.5 397B, and
GLM 5.2. Judges could evaluate hypotheses from their own developers because
excluding them would introduce unequal panel composition across arms, a greater
confound than potential within-developer preference. Including multiple
providers and open- and closed-weight models reduced dependence on individual
model preferences.

\subsection{Frontier model comparison}

Four frontier models available in February 2026 (GPT-5.2-high, Gemini 3.1 Pro
Preview, Claude Opus 4.6, and Grok 4.1 Fast Reasoning) were evaluated with
built-in web search enabled. Each model was instructed to generate up to 10
distinct, scientifically plausible, novel, feasible, and contributive hypotheses
per objective, returning none if no candidates met these criteria
(\snoteone). All models received the same 50 research objectives,
comprising goals and constraints. Gemini 3.1 Pro Preview, GPT-5.2-high, Claude
Opus 4.6, and Grok 4.1 Fast Reasoning generated 263, 418, 407, and 229
hypotheses, respectively. Claude Opus 4.6 returned no hypothesis for three
objectives, whereas the other models generated hypotheses for all 50 objectives.
The Ideation Core generated 293 hypotheses across 47 objectives, with none
surviving Refinement for the remaining three. Because Ideation Core hypotheses
differ from single-model abstracts in length and structure, potentially biasing
LLM judgments and embedding distances, GPT-5.5 summarized each into a
130--200-word abstract in plain academic language using a prompt matched to the
frontier models' output requirements. The summarizer performed no independent
retrieval or reasoning. These summaries, rather than the original hypotheses,
were evaluated using identical embedding, literature-retrieval, and judging
procedures across all models.

\subsection{Component ablation}

Each ablation bypassed one Ideation Core step while retaining the remaining
pipeline, using the full pipeline as reference. Formulation bypass replaced
generated formulations with repeated copies of the original objective,
preserving downstream execution counts while eliminating distinct framings.
Element Selection bypass omitted element discovery and recombination, generating
hypotheses directly from each formulation and objective using a modified
ideation prompt without element analysis or selection. Refinement bypass
retained hypotheses directly from ideation, without multi-agent debate,
revision, or filtering. All arms used identical objectives, upstream stages, and
a content-addressed cache keyed by stage inputs, recomputing only the bypassed
and downstream stages while reusing other outputs. For Formulation bypass, cache
keys additionally included instance indices to process all ten identical
objectives independently. A verification script confirmed that each arm matched
the full pipeline upstream of the bypass. The Formulation, Element Selection,
and Refinement bypass arms retained 388 (50 objectives), 287 (44 objectives),
and 376 (50 objectives) hypotheses, respectively. Ablations were evaluated using
raw Ideation Core outputs rather than summaries, as output formats were
identical across arms. Each step was assessed against its targeted property:
diversity (Formulation), novelty measured by literature distance (Element
Selection), and scientific soundness (Refinement). Because Refinement may
obscure upstream contributions by modifying every hypothesis, we also compared
literature distance before Refinement in the full and Element Selection-bypass
arms (\cref{fig:evaluation}e, left), using the same intermediate outputs
returned by Refinement bypass. All 29 hypotheses rejected by Refinement as
unsalvageable were excluded from analysis.

\subsection{Proposal generation for the intestinal barrier objective}

The investigator's initial goal was to develop a therapeutic strategy for
controlling age-associated intestinal epithelial leakage. Dialogue with the
Gateway LLM refined this into an objective with key constraints (exact input in
\snotetwo). The Ideation Core then generated nine reports in a
single uninterrupted run without human intervention, reruns, or manual filtering
of intermediate outputs. All raw outputs are publicly available (see
\nameref{sec:data-availability}).

The investigator evaluated the nine reports for conceptual novelty relative to
existing intestinal barrier literature, mechanistic plausibility, and
feasibility using available \textit{in vitro} models. One report presented a
particularly intriguing concept, prompting two follow-up objectives addressing
PEG-based surface modification and cholesterol-DNA nanobridges. Both specified
Caco-2 monolayers as the \textit{in vitro} barrier model and TEER as the
permeability readout (exact objectives in \snotetwo). Each
generated eight reports, from which the same investigator, along with two
additional investigators, selected one for experimental testing. All reports are
publicly available (see \nameref{sec:data-availability}), with the unmodified
cholesterol-DNA nanobridge report also provided in \sappone. Both
selected proposals were finalized and archived on 6 March 2026, before
experiments began on 1 April 2026.

\subsection{Cell culture and barrier model}

Caco-2 human intestinal epithelial cells (Korea Cell Line Bank, Seoul, Korea)
were seeded at $2 \times 10^{4}$ cells/well on polycarbonate Transwell inserts
(0.4 $\mu$m pores; SPL, Gyeonggi, Korea) and cultured for 21 days in Minimum
Essential Medium (Cytiva, Marlborough, MA, USA) containing 10\% fetal bovine
serum (Gibco, Waltham, MA, USA) and 1\% penicillin--streptomycin to ensure
differentiation and tight junction maturation. Medium was replaced every 2--3
days. Barrier disruption was induced with 20 mM M$\beta$CD (Sigma-Aldrich,
St.\ Louis, MO, USA) in serum-free medium for 1
h\autocite{mahammad2014cholesterol,lambert2005depletion}, to achieve
approximately 40\% TEER reduction. Monolayers were then washed 2--3 times with
phosphate-buffered saline (PBS) before further treatment.

\subsection{DSPE-PEG treatment and geometry comparison}

DSPE-PEG2000 (BroadPharm, San Diego, CA, USA) was applied apically (50, 100
$\mu$M), basolaterally (50, 100 $\mu$M), or to both surfaces (25, 50 $\mu$M per
side) using a dose-matched design. Free PEG2000 (100, 500 $\mu$M) was applied
apically as a non-tethered control. Dose matching ensured equivalent total PEG
exposure between single- and dual-surface treatments (e.g., 50 $\mu$M apically
vs.\ 25 $\mu$M per side). Following M$\beta$CD washout, treatments were
administered for 1 h at 37\,$^{\circ}$C. Each condition was tested in triplicate
across two independent experiments using different cell passages (six wells per
condition).

\subsection{DNA nanobridge assembly and toehold displacement}

DNA oligonucleotides with 5$'$-cholesterol-TEG modifications were synthesized by
Integrated DNA Technologies (IDT; Coralville, IA, USA): A,
5$'$-Chol-TEG-ATACGACT-3$'$ (8 nt); B, 5$'$-Chol-TEG-AGTCGTAT-3$'$ (8 nt,
reverse complement of A); and Bscr, 5$'$-Chol-TEG-TTAGCGAT-3$'$ (8 nt, shuffled
control with identical base composition). For toehold-mediated displacement,
Btoe (5$'$-Chol-TEG-tcagtcAGTCGTAT-3$'$, 14 nt) replaced B, with lowercase
letters indicating the 6-nt toehold. The unmodified invader
(5$'$-ATACGACTGACTGA-3$'$, 14 nt) was complementary to full-length Btoe. IDT
OligoAnalyzer confirmed minimal hairpin and self-dimer formation, and its
heterodimer tool predicted duplex binding free energy using default parameters.
For bridge assembly, M$\beta$CD-treated monolayers received apical A (0.5
$\mu$M) for 1 h, followed by washing and apical B, Bscr, or Btoe (0.5 $\mu$M)
for 1 h. For displacement, the invader (2.5--5 $\mu$M; 5--10-fold molar excess
over Btoe) was applied apically for 1 h after confirming TEER recovery.
Cholesterol--M$\beta$CD complex was administered alongside A as a positive
control for barrier recovery. Each condition was tested in triplicate across two
independent experiments using different cell passages (six wells per condition).

\subsection{TEER measurement}

TEER was measured using a Millicell ERS meter (Millipore, Bedford, MA, USA) at
baseline (0 h), after M$\beta$CD treatment (1 h), and hourly after each
subsequent treatment. Electrodes were rinsed with Dulbecco's PBS (DPBS) between
wells to prevent cross-contamination. Cell-free insert resistance was
subtracted, and TEER ($\Omega \cdot \mathrm{cm}^{2}$) was normalized to baseline
and expressed as a percentage of pretreatment values relative to controls.

\subsection{FITC--dextran paracellular flux assay}

Paracellular permeability was assessed using 4-kDa FITC--dextran
(Sigma-Aldrich). Following PEG or nanobridge treatment, monolayers were washed
with PBS, and FITC--dextran (1 mg/mL in DPBS) and fresh DPBS were added to the
apical and basolateral chambers, respectively. After 2 h at 37\,$^{\circ}$C
protected from light, basolateral fluorescence was measured at 485 nm excitation
and 535 nm emission (VICTOR X4, PerkinElmer, Waltham, MA, USA). Flux and TEER
were assessed in separate experiments using identical treatments and cell
preparations. Pilot comparisons of 1- and 2-h incubations under control and
M$\beta$CD conditions established the duration required for adequate
signal-to-noise ratio. Each condition was tested in duplicate across two
independent experiments using different cell passages (four wells per
condition).

\subsection{Transwell membrane immunofluorescence and confocal imaging}

Treated monolayers were fixed in 4\% paraformaldehyde for 15--20 min immediately
after treatment, washed with PBS, permeabilized with 0.1\% Triton X-100 for 10
min, and blocked with 10\% normal goat serum (Vector Laboratories, Newark, CA,
USA) for 1 h. Membranes were incubated with anti-ZO-1 polyclonal antibody
(1:50; Invitrogen, Waltham, MA, USA) overnight at 4\,$^{\circ}$C, followed by
Alexa Fluor 488-conjugated goat anti-rabbit IgG (H+L) secondary antibody
(1:500; Abcam, Cambridge, UK) for 1 h at room temperature. Rhodamine-DSPE-PEG
(BroadPharm) provided intrinsic red fluorescence without staining. Membranes
were excised using a \#11 scalpel blade and mounted cell-side down in ProLong
Gold Antifade containing 4$'$,6-diamidino-2-phenylindole (DAPI; Molecular
Probes, Eugene, OR, USA). Confocal images were acquired using an LSM900
microscope (Carl Zeiss, Oberkochen, Germany) with a 63$\times$ oil-immersion
objective (NA $\ge$ 1.4) and sequential scanning. All conditions were imaged in
one session using identical laser power, gain and pinhole settings. Z-stacks
comprised 130 slices at 0.25 $\mu$m intervals (32.25 $\mu$m total) over a
$77 \times 77$ $\mu$m field. XZ orthogonal reconstructions visualized
rhodamine-DSPE-PEG and ZO-1 distribution across apical, lateral, and basal
membrane domains. Three fields per well were imaged from two wells per condition
in each of two independent experiments. Rhodamine intensity profiles were
generated in Fiji from XZ reconstructions using perpendicular lines at ten
positions per image, with one image per condition from each independent
experiment (20 profiles/condition). Each profile spanned the DAPI-defined
apical-to-basal extent and was sampled at 200 points over 0--25.2 $\mu$m.
Background-subtracted raw fluorescence intensities were averaged across 20
profiles per condition.

\subsection{Statistical analysis}

\textbf{System evaluation and ablation.} The objective served as the unit of
analysis. Literature distance and soundness were averaged across hypotheses per
objective, whereas inter-hypothesis diversity was calculated for objectives with
$\ge$ 2 hypotheses. Arms were compared using two-sided paired Wilcoxon
signed-rank tests on shared objectives, with effect sizes expressed as median
within-objective differences. $p$ values were Holm-adjusted across four frontier
models per measure (\cref{fig:evaluation}c) and three ablated stages
(\cref{fig:evaluation}d--f). The pre-Refinement Element Selection comparison
(\cref{fig:evaluation}e, left) was unadjusted. In \cref{fig:evaluation}g,
differences were standardized by the full pipeline's across-objective standard
deviation, with the six comparisons absent from \cref{fig:evaluation}d--f
Holm-adjusted separately. \Cref{fig:evaluation}a,b are descriptive; literature
distance and soundness were z-scored across all hypotheses, and
\cref{fig:evaluation}b shows 95\% bootstrap intervals across objectives (1,000
resamples). Analyses used Python 3.14.4, NumPy 2.5.1, and SciPy 1.18.0.

\textbf{Wet-lab experiments.} Wells were the units of analysis and were pooled
across two independent experiments per assay. Group differences were assessed by
one-way ANOVA with Dunnett's (versus M$\beta$CD), Tukey's (all pairs), or
Bonferroni's (prespecified comparisons) multiple-comparisons tests, as specified
in figure legends. Two-factor comparisons used two-way repeated-measures ANOVA
followed by Dunnett's or Tukey's tests. Statistical significance was set at
$p < 0.05$, with tests selected according to data normality. Analyses used
GraphPad Prism 10.3.1 (GraphPad, San Diego, CA, USA). Investigators were not
blinded to group allocation.

% ==========================================================
% Back Matter
% ==========================================================

% ============== Back Matter Section ============== %

\section{Data availability}
\label{sec:data-availability}

The quantitative evaluation datasets, including input research objectives,
generated hypotheses, and evaluation results, and Abelio-generated proposals
selected for wet-lab validation are publicly available at
\url{https://github.com/Asteromorph-Corp/abelio-1.0}. Wet-lab experimental
results are presented in this article and its
\hyperref[sec:supplementary]{Supplementary Information}.

\section{Code availability}
\label{sec:code-availability}

The Abelio source code, including the Ideation Core and Report Writer, and
quantitative evaluation scripts are publicly available at
\url{https://github.com/Asteromorph-Corp/abelio-1.0}.

\clearpage

{\hypersetup{urlcolor=black}
  \printbibliography[title={References}]

@article{senior2020improved,
  title={Improved protein structure prediction using potentials from deep learning},
  author={Senior, Andrew W and Evans, Richard and Jumper, John and Kirkpatrick, James and Sifre, Laurent and Green, Tim and Qin, Chongli and {\v{Z}}{\'\i}dek, Augustin and Nelson, Alexander WR and Bridgland, Alex and others},
  journal={Nature},
  volume={577},
  number={7792},
  pages={706--710},
  year={2020},
  publisher={Nature Publishing Group UK London}
}

@article{jumper2021highly,
  title={Highly accurate protein structure prediction with AlphaFold},
  author={Jumper, John and Evans, Richard and Pritzel, Alexander and Green, Tim and Figurnov, Michael and Ronneberger, Olaf and Tunyasuvunakool, Kathryn and Bates, Russ and {\v{Z}}{\'\i}dek, Augustin and Potapenko, Anna and others},
  journal={nature},
  volume={596},
  number={7873},
  pages={583--589},
  year={2021},
  publisher={Nature Publishing Group UK London}
}

@article{abramson2024accurate,
  title={Accurate structure prediction of biomolecular interactions with AlphaFold 3},
  author={Abramson, Josh and Adler, Jonas and Dunger, Jack and Evans, Richard and Green, Tim and Pritzel, Alexander and Ronneberger, Olaf and Willmore, Lindsay and Ballard, Andrew J and Bambrick, Joshua and others},
  journal={Nature},
  volume={630},
  number={8016},
  pages={493--500},
  year={2024},
  publisher={Nature Publishing Group UK London}
}

@article{watson2023novo,
  title={De novo design of protein structure and function with RFdiffusion},
  author={Watson, Joseph L and Juergens, David and Bennett, Nathaniel R and Trippe, Brian L and Yim, Jason and Eisenach, Helen E and Ahern, Woody and Borst, Andrew J and Ragotte, Robert J and Milles, Lukas F and others},
  journal={Nature},
  volume={620},
  number={7976},
  pages={1089--1100},
  year={2023},
  publisher={Nature Publishing Group UK London}
}

@article{lam2023learning,
  title={Learning skillful medium-range global weather forecasting},
  author={Lam, Remi and Sanchez-Gonzalez, Alvaro and Willson, Matthew and Wirnsberger, Peter and Fortunato, Meire and Alet, Ferran and Ravuri, Suman and Ewalds, Timo and Eaton-Rosen, Zach and Hu, Weihua and others},
  journal={Science},
  volume={382},
  number={6677},
  pages={1416--1421},
  year={2023},
  publisher={American Association for the Advancement of Science}
}

@article{merchant2023scaling,
  title={Scaling deep learning for materials discovery},
  author={Merchant, Amil and Batzner, Simon and Schoenholz, Samuel S and Aykol, Muratahan and Cheon, Gowoon and Cubuk, Ekin Dogus},
  journal={Nature},
  volume={624},
  number={7990},
  pages={80--85},
  year={2023},
  publisher={Nature Publishing Group UK London}
}

@article{degrave2022magnetic,
  title={Magnetic control of tokamak plasmas through deep reinforcement learning},
  author={Degrave, Jonas and Felici, Federico and Buchli, Jonas and Neunert, Michael and Tracey, Brendan and Carpanese, Francesco and Ewalds, Timo and Hafner, Roland and Abdolmaleki, Abbas and de Las Casas, Diego and others},
  journal={Nature},
  volume={602},
  number={7897},
  pages={414--419},
  year={2022},
  publisher={Nature Publishing Group UK London}
}

@article{liu2023deep,
  title={Deep learning-guided discovery of an antibiotic targeting Acinetobacter baumannii},
  author={Liu, Gary and Catacutan, Denise B and Rathod, Khushi and Swanson, Kyle and Jin, Wengong and Mohammed, Jody C and Chiappino-Pepe, Anush and Syed, Saad A and Fragis, Meghan and Rachwalski, Kenneth and others},
  journal={Nature Chemical Biology},
  volume={19},
  number={11},
  pages={1342--1350},
  year={2023},
  publisher={Nature Publishing Group US New York}
}

@article{wang2023scientific,
  title={Scientific discovery in the age of artificial intelligence},
  author={Wang, Hanchen and Fu, Tianfan and Du, Yuanqi and Gao, Wenhao and Huang, Kexin and Liu, Ziming and Chandak, Payal and Liu, Shengchao and Van Katwyk, Peter and Deac, Andreea and others},
  journal={Nature},
  volume={620},
  number={7972},
  pages={47--60},
  year={2023},
  publisher={Nature Publishing Group UK London}
}

@article{gao2024empowering,
  title={Empowering biomedical discovery with AI agents},
  author={Gao, Shanghua and Fang, Ada and Huang, Yepeng and Giunchiglia, Valentina and Noori, Ayush and Schwarz, Jonathan Richard and Ektefaie, Yasha and Kondic, Jovana and Zitnik, Marinka},
  journal={Cell},
  volume={187},
  number={22},
  pages={6125--6151},
  year={2024},
  publisher={Elsevier}
}

@misc{ren2025towards,
      title={Towards Scientific Intelligence: A Survey of LLM-based Scientific Agents}, 
      author={Shuo Ren and Can Xie and Pu Jian and Zhenjiang Ren and Chunlin Leng and Jiajun Zhang},
      year={2025},
      eprint={2503.24047},
      archivePrefix={arXiv},
      primaryClass={cs.AI},
      url={https://arxiv.org/abs/2503.24047}, 
}

@article{zhang2025exploring,
  title={Exploring the role of large language models in the scientific method: from hypothesis to discovery},
  author={Zhang, Yanbo and Khan, Sumeer A and Mahmud, Adnan and Yang, Huck and Lavin, Alexander and Levin, Michael and Frey, Jeremy and Dunnmon, Jared and Evans, James and Bundy, Alan and others},
  journal={npj Artificial Intelligence},
  volume={1},
  number={1},
  pages={14},
  year={2025},
  publisher={Nature Publishing Group UK London}
}

@article{boiko2023autonomous,
  title={Autonomous chemical research with large language models},
  author={Boiko, Daniil A and MacKnight, Robert and Kline, Ben and Gomes, Gabe},
  journal={Nature},
  volume={624},
  number={7992},
  pages={570--578},
  year={2023},
  publisher={Nature Publishing Group UK London}
}

@article{ifargan2025autonomous,
  title={Autonomous LLM-driven research—from data to human-verifiable research papers},
  author={Ifargan, Tal and Hafner, Lukas and Kern, Maor and Alcalay, Ori and Kishony, Roy},
  journal={NEJM AI},
  volume={2},
  number={1},
  pages={AIoa2400555},
  year={2025},
  publisher={Massachusetts Medical Society}
}

@article{lu2026towards,
  title={Towards end-to-end automation of AI research},
  author={Lu, Chris and Lu, Cong and Lange, Robert Tjarko and Yamada, Yutaro and Hu, Shengran and Foerster, Jakob and Ha, David and Clune, Jeff},
  journal={Nature},
  volume={651},
  number={8107},
  pages={914--919},
  year={2026},
  publisher={Nature Publishing Group UK London}
}

@article{gottweis2026accelerating,
  title={Accelerating scientific discovery with Co-Scientist},
  volume={655},
  ISSN={1476-4687},
  url={http://dx.doi.org/10.1038/s41586-026-10644-y},
  DOI={10.1038/s41586-026-10644-y},
  number={8122},
  journal={Nature},
  publisher={Springer Science and Business Media LLC},
  author={Gottweis, Juraj and Weng, Wei-Hung and Daryin, Alexander and Tu, Tao and Sirkovic, Petar and Myaskovsky, Artiom and Glowaty, Grzegorz and Weissenberger, Felix and Orlandi, Alessio and Popovici, Dan and Palepu, Anil and Rong, Keran and Tanno, Ryutaro and Saab, Khaled and Zhang, Fan and Blum, Jacob and Carroll, Andrew and Kulkarni, Kavita and Tomašev, Nenad and Zverinski, Dina and Rendulic, Ivor and Vedadi, Elahe and Hasler, Florian and Rimanic, Luka and Boia, Marina and Budiselic, Ivan and Feinstein, Ben and Bellaiche, Mathias and Sheffer, Tom and Freyberg, Jan and Ratcliff, Jeremy and Bertolli, Ottavia and Chou, Katherine and Hassidim, Avinatan and Gokturk, Burak and Vahdat, Amin and Guan, Yuan and Dhillon, Vikram and Vaishnav, Eeshit Dhaval and Lee, Byron and Costa, Tiago R. D. and Penadés, José R. and Peltz, Gary and Matias, Yossi and Manyika, James and Hassabis, Demis and Xu, Yunhan and Kohli, Pushmeet and Pawlosky, Annalisa and Karthikesalingam, Alan and Natarajan, Vivek},
  year={2026},

  pages={487--496}
}

@article{ghareeb2026multi,
  title={A multi-agent system for automating scientific discovery},
  volume={655},
  ISSN={1476-4687},
  url={http://dx.doi.org/10.1038/s41586-026-10652-y},
  DOI={10.1038/s41586-026-10652-y},
  number={8122},
  journal={Nature},
  publisher={Springer Science and Business Media LLC},
  author={Ghareeb, Ali E. and Chang, Benjamin and Mitchener, Ludovico and Yiu, Angela and Szostkiewicz, Caralyn J. and Shved, Dmytro and Gyimesi, Gavin J. and Laurent, Jon M. and Wright, Samantha M. and Razzak, Muhammed T. and White, Andrew D. and Finnemann, Silvia C. and Hinks, Michaela M. and Rodriques, Samuel G.},
  year={2026},

  pages={497--505}
}

@article{boyden2005millisecond,
  title={Millisecond-timescale, genetically targeted optical control of neural activity},
  author={Boyden, Edward S and Zhang, Feng and Bamberg, Ernst and Nagel, Georg and Deisseroth, Karl},
  journal={Nature neuroscience},
  volume={8},
  number={9},
  pages={1263--1268},
  year={2005},
  publisher={Nature Publishing Group US New York}
}

@article{saxon2000cell,
  title={Cell surface engineering by a modified Staudinger reaction},
  author={Saxon, Eliana and Bertozzi, Carolyn R},
  journal={Science},
  volume={287},
  number={5460},
  pages={2007--2010},
  year={2000},
  publisher={American Association for the Advancement of Science}
}

@article{kaplan2025global,
  title={The global burden of inflammatory bowel disease: from 2025 to 2045},
  author={Kaplan, Gilaad G},
  journal={Nature Reviews Gastroenterology \& Hepatology},
  volume={22},
  number={10},
  pages={708--720},
  year={2025},
  publisher={Nature Publishing Group UK London}
}

@article{singh2018global,
  title={Global prevalence of celiac disease: systematic review and meta-analysis},
  author={Singh, Prashant and Arora, Ananya and Strand, Tor A and Leffler, Daniel A and Catassi, Carlo and Green, Peter H and Kelly, Ciaran P and Ahuja, Vineet and Makharia, Govind K},
  journal={Clinical gastroenterology and hepatology},
  volume={16},
  number={6},
  pages={823--836},
  year={2018},
  publisher={Elsevier}
}

@article{odenwald2017intestinal,
  title={The intestinal epithelial barrier: a therapeutic target?},
  author={Odenwald, Matthew A and Turner, Jerrold R},
  journal={Nature reviews Gastroenterology \& hepatology},
  volume={14},
  number={1},
  pages={9--21},
  year={2017},
  publisher={Nature Publishing Group UK London}
}

@article{camilleri2019leaky,
  title={Leaky gut: mechanisms, measurement and clinical implications in humans},
  author={Camilleri, Michael},
  journal={Gut},
  volume={68},
  number={8},
  pages={1516--1526},
  year={2019},
  publisher={BMJ Publishing Group}
}

@article{moonwiriyakit2023tight,
  title={Tight junctions: from molecules to gastrointestinal diseases},
  author={Moonwiriyakit, Aekkacha and Pathomthongtaweechai, Nutthapoom and Steinhagen, Peter R and Chantawichitwong, Papasara and Satianrapapong, Wilasinee and Pongkorpsakol, Pawin},
  journal={Tissue Barriers},
  volume={11},
  number={2},
  pages={2077620},
  year={2023},
  publisher={Taylor \& Francis}
}

@article{horowitz2023paracellular,
  title={Paracellular permeability and tight junction regulation in gut health and disease},
  author={Horowitz, Arie and Chanez-Paredes, Sandra D and Haest, Xenia and Turner, Jerrold R},
  journal={Nature Reviews Gastroenterology \& Hepatology},
  volume={20},
  number={7},
  pages={417--432},
  year={2023},
  publisher={Nature Publishing Group UK London}
}

@article{ahmad2017gut,
  title={Gut permeability and mucosal inflammation: bad, good or context dependent},
  author={Ahmad, R and Sorrell, MF and Batra, SK and Dhawan, P and Singh, AB},
  journal={Mucosal immunology},
  volume={10},
  number={2},
  pages={307--317},
  year={2017},
  publisher={Elsevier}
}

@article{schumann2017celiac,
  title={Celiac disease: role of the epithelial barrier},
  author={Schumann, Michael and Siegmund, Britta and Schulzke, J{\"o}rg D and Fromm, Michael},
  journal={Cellular and molecular gastroenterology and hepatology},
  volume={3},
  number={2},
  pages={150--162},
  year={2017},
  publisher={Elsevier}
}

@article{pentinmikko2019notum,
  title={Notum produced by Paneth cells attenuates regeneration of aged intestinal epithelium},
  author={Pentinmikko, Nalle and Iqbal, Sharif and Mana, Miyeko and Andersson, Simon and Cognetta III, Armand B and Suciu, Radu M and Roper, Jatin and Luopaj{\"a}rvi, Kalle and Markelin, Eino and Gopalakrishnan, Swetha and others},
  journal={Nature},
  volume={571},
  number={7765},
  pages={398--402},
  year={2019},
  publisher={Nature Publishing Group UK London}
}

@article{omrani2023ifngamma,
  title={IFN$\gamma$-Stat1 axis drives aging-associated loss of intestinal tissue homeostasis and regeneration},
  author={Omrani, Omid and Krepelova, Anna and Rasa, Seyed Mohammad Mahdi and Sirvinskas, Dovydas and Lu, Jing and Annunziata, Francesco and Garside, George and Bajwa, Seerat and Reinhardt, Susanne and Adam, Lisa and others},
  journal={Nature communications},
  volume={14},
  number={1},
  pages={6109},
  year={2023},
  publisher={Nature Publishing Group UK London}
}

@article{graham2019intracellular,
  title={Intracellular MLCK1 diversion reverses barrier loss to restore mucosal homeostasis},
  author={Graham, W Vallen and He, Weiqi and Marchiando, Amanda M and Zha, Juanmin and Singh, Gurminder and Li, Hua-Shan and Biswas, Amlan and Ong, Ma Lora Drizella M and Jiang, Zhi-Hui and Choi, Wangsun and others},
  journal={Nature medicine},
  volume={25},
  number={4},
  pages={690--700},
  year={2019},
  publisher={Nature Publishing Group US New York}
}

@article{schuppan2021randomized,
  title={A randomized trial of a transglutaminase 2 inhibitor for celiac disease},
  author={Schuppan, Detlef and M{\"a}ki, Markku and Lundin, Knut EA and Isola, Jorma and Friesing-Sosnik, Tina and Taavela, Juha and Popp, Alina and Koskenpato, Jari and Langhorst, Jost and Hovde, {\O}istein and others},
  journal={New England Journal of Medicine},
  volume={385},
  number={1},
  pages={35--45},
  year={2021},
  publisher={Mass Medical Soc}
}

@article{ghosh2021regulation,
  title={Regulation of intestinal barrier function by microbial metabolites},
  author={Ghosh, Sweta and Whitley, Caleb Samuel and Haribabu, Bodduluri and Jala, Venkatakrishna Rao},
  journal={Cellular and molecular gastroenterology and hepatology},
  volume={11},
  number={5},
  pages={1463--1482},
  year={2021},
  publisher={Elsevier}
}

@article{cropley2006praise,
  title={In praise of convergent thinking},
  author={Cropley, Arthur},
  journal={Creativity research journal},
  volume={18},
  number={3},
  pages={391--404},
  year={2006},
  publisher={Taylor \& Francis}
}

@article{dym2005engineering,
  title={Engineering design thinking, teaching, and learning},
  author={Dym, Clive L and Agogino, Alice M and Eris, Ozgur and Frey, Daniel D and Leifer, Larry J},
  journal={Journal of engineering education},
  volume={94},
  number={1},
  pages={103--120},
  year={2005},
  publisher={Wiley Online Library}
}

@misc{sinhahajari2026limitsllmasjudgescientificnovelty,
      title={On the Limits of LLM-as-Judge for Scientific Novelty Assessment}, 
      author={Soumitra Sinhahajari and Navonil Majumder and Soujanya Poria},
      year={2026},
      eprint={2606.12071},
      archivePrefix={arXiv},
      primaryClass={cs.DL},
      url={https://arxiv.org/abs/2606.12071}, 
}

@misc{baumann2026stopautomatingpeerreview,
      title={Stop Automating Peer Review Without Rigorous Evaluation}, 
      author={Joachim Baumann and Jiaxin Pei and Sanmi Koyejo and Dirk Hovy},
      year={2026},
      eprint={2605.03202},
      archivePrefix={arXiv},
      primaryClass={cs.AI},
      url={https://arxiv.org/abs/2605.03202}, 
}

@article{ruan2026evaluating,
  title={Evaluating LLMs' divergent thinking capabilities for scientific idea generation with minimal context},
  author={Ruan, Kai and Wang, Xuan and Hong, Jixiang and Wang, Peng and Liu, Yang and Sun, Hao},
  journal={Nature communications},
  volume={17},
  number={1},
  pages={3625},
  year={2026},
  publisher={Nature Publishing Group UK London}
}

@article{wu2004high,
  title={High-molecular-weight polyethylene glycol prevents lethal sepsis due to intestinal Pseudomonas aeruginosa},
  author={Wu, Licheng and Zaborina, Olga and Zaborin, Alex and Chang, Eugene B and Musch, Mark and Holbrook, Christopher and Shapiro, James and Turner, Jerrold R and Wu, Guohui and Alverdy, John C and others},
  journal={Gastroenterology},
  volume={126},
  number={2},
  pages={488--498},
  year={2004},
  publisher={Elsevier}
}

@article{valuckaite2009oral,
  title={Oral PEG 15--20 protects the intestine against radiation: role of lipid rafts},
  author={Valuckaite, Vesta and Zaborina, Olga and Long, Jason and Hauer-Jensen, Martin and Wang, Junru and Holbrook, Christopher and Zaborin, Alexander and Drabik, Kenneth and Katdare, Mukta and Mauceri, Helena and others},
  journal={American Journal of Physiology-Gastrointestinal and Liver Physiology},
  volume={297},
  number={6},
  pages={G1041--G1052},
  year={2009},
  publisher={American Physiological Society Bethesda, MD}
}

@article{bharadwaj2011higher,
  % Only the \emph{} markup was added here; the wording is as retrieved.
  title={Higher molecular weight polyethylene glycol increases cell proliferation while improving barrier function in an \emph{in vitro} colon cancer model},
  author={Bharadwaj, Shruthi and Vishnubhotla, Ramana and Shan, Sun and Chauhan, Chinmay and Cho, Michael and Glover, Sarah C},
  journal={Journal of Biomedicine and Biotechnology},
  volume={2011},
  number={1},
  pages={587470},
  year={2011},
  publisher={Wiley Online Library}
}

@article{casselbrant2018luminal,
  title={Luminal polyethylene glycol alleviates intestinal preservation injury irrespective of molecular size},
  author={Casselbrant, Anna and S{\"o}fteland, John M and Hellstr{\"o}m, Mats and Malinauskas, Mantas and Oltean, Mihai},
  journal={The Journal of Pharmacology and Experimental Therapeutics},
  volume={366},
  number={1},
  pages={29--36},
  year={2018},
  publisher={Elsevier}
}

@article{gao2023dspe,
  title={DSPE-PEG polymer enhanced berberine absorption specifically in the small intestine of rats through paracellular passway},
  author={Gao, Yang and Sun, Jianmei and Wang, Shucong and Huxiao, Lisong and Xu, Yali and Zhang, Hailong},
  journal={Journal of Pharmacy and Pharmacology},
  volume={75},
  number={7},
  pages={931--939},
  year={2023},
  publisher={Oxford University Press UK}
}

@article{lambert2005depletion,
  title={Depletion of Caco-2 cell cholesterol disrupts barrier function by altering the detergent solubility and distribution of specific tight-junction proteins},
  author={Lambert, Daniel and O'NEILL, Catherine A and Padfield, Philip J},
  journal={Biochemical Journal},
  volume={387},
  number={2},
  pages={553--560},
  year={2005},
  publisher={Portland Press Ltd.}
}

@article{mahammad2014cholesterol,
  title={Cholesterol depletion using methyl-$\beta$-cyclodextrin},
  author={Mahammad, Saleemulla and Parmryd, Ingela},
  journal={Methods in Molecular Biology},
  volume={1232},
  pages={91--102},
  year={2015},
  publisher={Springer}
}

@article{zhang2009control,
  title={Control of DNA strand displacement kinetics using toehold exchange},
  author={Zhang, David Yu and Winfree, Erik},
  journal={Journal of the American Chemical Society},
  volume={131},
  number={47},
  pages={17303--17314},
  year={2009},
  publisher={ACS Publications}
}

@article{seeman2017dna,
  title={DNA nanotechnology},
  author={Seeman, Nadrian C and Sleiman, Hanadi F},
  journal={Nature Reviews Materials},
  volume={3},
  number={1},
  pages={17068},
  year={2018},
  publisher={Nature Publishing Group}
}

@article{pinheiro2011challenges,
  title={Challenges and opportunities for structural DNA nanotechnology},
  author={Pinheiro, Andre V and Han, Dongran and Shih, William M and Yan, Hao},
  journal={Nature nanotechnology},
  volume={6},
  number={12},
  pages={763--772},
  year={2011},
  publisher={Nature Publishing Group UK London}
}

@article{schoenit2021functionalization,
  title={Functionalization of cellular membranes with DNA nanotechnology},
  author={Schoenit, Andreas and Cavalcanti-Adam, Elisabetta Ada and G{\"o}pfrich, Kerstin},
  journal={Trends in biotechnology},
  volume={39},
  number={11},
  pages={1208--1220},
  year={2021},
  publisher={Elsevier}
}

@article{you2017dna,
  title={DNA probes for monitoring dynamic and transient molecular encounters on live cell membranes},
  author={You, Mingxu and Lyu, Yifan and Han, Da and Qiu, Liping and Liu, Qiaoling and Chen, Tao and Sam Wu, Cuichen and Peng, Lu and Zhang, Liqin and Bao, Gang and others},
  journal={Nature nanotechnology},
  volume={12},
  number={5},
  pages={453--459},
  year={2017},
  publisher={Nature Publishing Group UK London}
}

@article{gartner2009programmed,
  title={Programmed assembly of 3-dimensional microtissues with defined cellular connectivity},
  author={Gartner, Zev J and Bertozzi, Carolyn R},
  journal={Proceedings of the National Academy of Sciences},
  volume={106},
  number={12},
  pages={4606--4610},
  year={2009},
  publisher={National Academy of Sciences}
}

@article{ge2020programming,
  title={Programming cell--cell communications with engineered cell origami clusters},
  author={Ge, Zhilei and Liu, Jiangbo and Guo, Linjie and Yao, Guangbao and Li, Qian and Wang, Lihua and Li, Jiang and Fan, Chunhai},
  journal={Journal of the American Chemical Society},
  volume={142},
  number={19},
  pages={8800--8808},
  year={2020},
  publisher={ACS Publications}
}

@article{kim2026amphiphilic,
  title={Amphiphilic Lipid--Single-Stranded DNA Conjugate-Mediated Cell Surface Engineering for Programmable Intercellular Tethering and Immune Synapse Formation},
  author={Kim, Sungjun and Lee, Chae Eun and Jangid, Ashok Kumar and Kim, Kyobum},
  journal={Biomaterials Research},
  volume={30},
  pages={0366},
  year={2026},
  publisher={AAAS}
}

@article{hoffecker2019tuning,
  title={Tuning intercellular adhesion with membrane-anchored oligonucleotides},
  author={Hoffecker, Ian T and Arima, Yusuke and Iwata, Hiroo},
  journal={Journal of The Royal Society Interface},
  volume={16},
  number={159},
  pages={20190299},
  year={2019}
}

@article{schoenit2021tuning,
  title={Tuning epithelial cell--cell adhesion and collective dynamics with functional DNA-E-cadherin hybrid linkers},
  author={Schoenit, Andreas and Lo Giudice, Cristina and Hahnen, Nina and Ollech, Dirk and Jahnke, Kevin and Göpfrich, Kerstin and Cavalcanti-Adam, Elisabetta Ada},
  journal={Nano letters},
  volume={22},
  number={1},
  pages={302--310},
  year={2022},
  publisher={ACS Publications}
}

@misc{sothanaphan2026resolution,
      title={Resolution of Erd\H{o}s Problem \#728: a writeup of Aristotle's Lean proof}, 
      author={Nat Sothanaphan},
      year={2026},
      eprint={2601.07421},
      archivePrefix={arXiv},
      primaryClass={math.NT},
      url={https://arxiv.org/abs/2601.07421}, 
}

@misc{tsoukalas2026advancing,
      title={Advancing Mathematics Research with AI-Driven Formal Proof Search}, 
      author={George Tsoukalas and Anton Kovsharov and Sergey Shirobokov and Anja Surina and Moritz Firsching and Gergely Bérczi and Francisco J. R. Ruiz and Arun Suggala and Adam Zsolt Wagner and Eric Wieser and Lei Yu and Aja Huang and Miklós Z. Horváth and Andrew Ferraiuolo and Henryk Michalewski and Edward Lockhart and Codrut Grosu and Thomas Hubert and Matej Balog and Pushmeet Kohli and Swarat Chaudhuri},
      year={2026},
      eprint={2605.22763},
      archivePrefix={arXiv},
      primaryClass={cs.AI},
      url={https://arxiv.org/abs/2605.22763}, 
}

@misc{jiang2025artificial,
      title={Artificial Hivemind: The Open-Ended Homogeneity of Language Models (and Beyond)}, 
      author={Liwei Jiang and Yuanjun Chai and Margaret Li and Mickel Liu and Raymond Fok and Nouha Dziri and Yulia Tsvetkov and Maarten Sap and Alon Albalak and Yejin Choi},
      year={2025},
      eprint={2510.22954},
      archivePrefix={arXiv},
      primaryClass={cs.CL},
      url={https://arxiv.org/abs/2510.22954}, 
}

@misc{priem2022openalex,
      title={OpenAlex: A fully-open index of scholarly works, authors, venues, institutions, and concepts}, 
      author={Jason Priem and Heather Piwowar and Richard Orr},
      year={2022},
      eprint={2205.01833},
      archivePrefix={arXiv},
      primaryClass={cs.DL},
      url={https://arxiv.org/abs/2205.01833}, 
}

@article{nagel2003channelrhodopsin,
  title={Channelrhodopsin-2, a directly light-gated cation-selective membrane channel},
  volume={100},
  ISSN={1091-6490},
  url={http://dx.doi.org/10.1073/pnas.1936192100},
  DOI={10.1073/pnas.1936192100},
  number={24},
  journal={Proceedings of the National Academy of Sciences},
  publisher={National Academy of Sciences},
  author={Nagel, Georg and Szellas, Tanjef and Huhn, Wolfram and Kateriya, Suneel and Adeishvili, Nona and Berthold, Peter and Ollig, Doris and Hegemann, Peter and Bamberg, Ernst},
  year={2003},

  pages={13940--13945}
}

@article{shimomura1962extraction,
  title={Extraction, Purification and Properties of Aequorin, a Bioluminescent Protein from the Luminous Hydromedusan, \textit{Aequorea}},
  volume={59},
  ISSN={1553-0809},
  url={http://dx.doi.org/10.1002/jcp.1030590302},
  DOI={10.1002/jcp.1030590302},
  number={3},
  journal={Journal of Cellular and Comparative Physiology},
  publisher={Wiley},
  author={Shimomura, Osamu and Johnson, Frank H. and Saiga, Yo},
  year={1962},

  pages={223--239}
}

@article{chalfie1994green,
  title={Green Fluorescent Protein as a Marker for Gene Expression},
  volume={263},
  ISSN={1095-9203},
  url={http://dx.doi.org/10.1126/science.8303295},
  DOI={10.1126/science.8303295},
  number={5148},
  journal={Science},
  publisher={American Association for the Advancement of Science (AAAS)},
  author={Chalfie, Martin and Tu, Yuan and Euskirchen, Ghia and Ward, William W. and Prasher, Douglas C.},
  year={1994},

  pages={802--805}
}

@online{openai2026ten,
  author={{OpenAI}},
  title={Ten Advances in Mathematics and Theoretical Computer Science},
  year={2026},
  url={https://openai.com/index/ten-advances-in-mathematics/},
  addendum={Lean certificates: \url{https://github.com/openai/ten-proofs}}
}

@online{anthropic2026riemann,
  author={{Anthropic}},
  title={Learning more about Claude's mathematical capabilities},
  year={2026},
  url={https://www.anthropic.com/research/riemann-zeta},
  addendum={Lean formalization: \url{https://github.com/anthropics/zeta-23-lean}}
}

@online{openai2026quasi,
  author={{OpenAI}},
  title={The Quasi-Riemann Hypothesis},
  year={2026},
  url={https://github.com/openai/math},
  addendum={Manuscript and Lean formalization, released October 2026}
}
}

% ==========================================================
% Supplementary Information
% ==========================================================

\clearpage

\appendix

% Supplementary figures are numbered from 1 again and labelled
% "Supplementary Fig. n." as in the Supplementary Information. In-text
% mentions use \sfigref, which prints the name explicitly, because the
% figures are cited in the main text before this renaming takes effect.
\setcounter{figure}{0}
\renewcommand{\figurename}{Supplementary Fig.}

% ============== Supplementary Information Section ============== %

\section{Supplementary Information}
\label{sec:supplementary}

\subsection{Supplementary Note 1. Prompts used for system evaluation}
\label{snote:prompts}

The following prompts were used for frontier-model comparison, scientific
soundness evaluation, and hypothesis summarization. Template variables
substituted at runtime are enclosed in double curly brackets.

\promptlisting{Prompt for frontier model comparison}%
  {resources/prompts/prompt_frontier.txt}

\promptlisting{Prompt for scientific soundness evaluation}%
  {resources/prompts/prompt_soundness.txt}

\promptlisting{Prompt for hypothesis summarization}%
  {resources/prompts/prompt_summarization.txt}

\subsection[Supplementary Note 2. Research objectives]{Supplementary Note 2.
  Research objectives used for prospective experimental validation}
\label{snote:objectives}

The exact research objectives and constraints submitted to Abelio across three
sequential runs for prospective experimental validation are provided below. The
number of proposals generated per run is specified in the corresponding box
title.

\promptlisting{Initial intestinal-barrier run (9 proposals)}%
  {resources/prompts/objective_initial.txt}

\promptlisting{Macromolecular-crowding follow-up run (8 proposals)}%
  {resources/prompts/objective_crowding.txt}

\promptlisting{Cholesterol-DNA nanobridge follow-up run (8 proposals)}%
  {resources/prompts/objective_nanobridge.txt}

\clearpage
\subsection{Supplementary Figures}
\label{sfig:figures}

\begin{figure}[!ht]
  \centering
  \includegraphics[width=\linewidth]{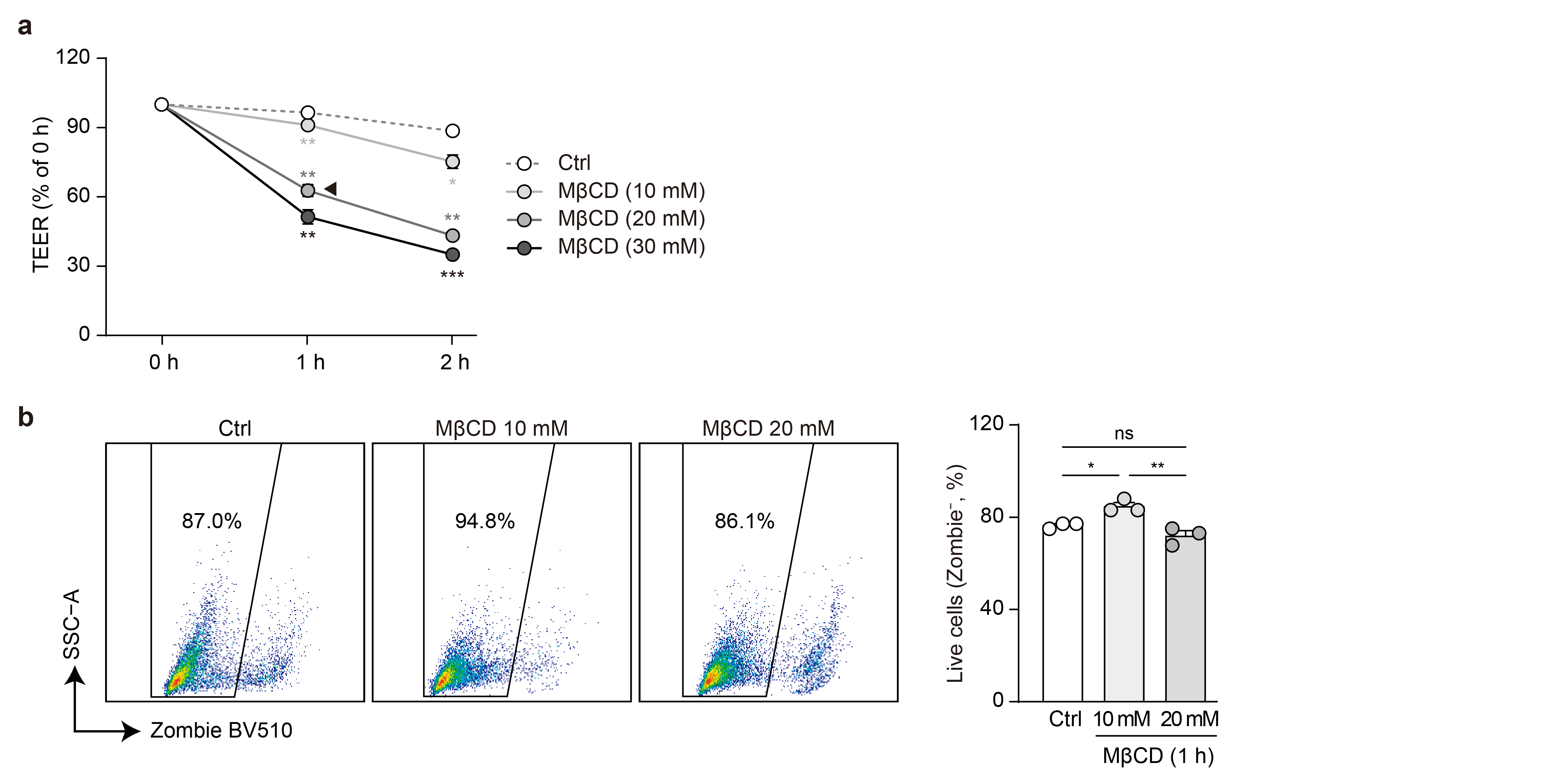}
  \caption[Optimization of M$\beta$CD treatment]{%
    \textbf{Optimization of M$\beta$CD treatment to establish a partially
    compromised Caco-2 barrier model.}
    \textbf{a}, Dose--response analysis of transepithelial electrical resistance
    (TEER) following methyl-$\beta$-cyclodextrin (M$\beta$CD) treatment.
    Color-matched asterisks indicate comparisons with the M$\beta$CD-only
    condition. The arrowhead indicates the condition selected for subsequent
    experiments (20 mM, 1 h).
    \textbf{b}, Cell viability under corresponding treatment conditions.
    Viability was assessed by flow cytometry of dissociated monolayers stained
    with Zombie Aqua (BioLegend, San Diego, CA, USA; 15 min at room temperature;
    BV510 channel), with Zombie Aqua-negative cells classified as viable.
    $^{*}p < 0.05$, $^{**}p < 0.01$, and $^{***}p < 0.001$ by two-way
    repeated-measures ANOVA with Dunnett's multiple comparisons test
    (\textbf{a}) or one-way ANOVA with Tukey's multiple comparisons test
    (\textbf{b}).%
  }
  \label{sfig:mbcd}
\end{figure}

\begin{figure}[!t]
  \centering
  \includegraphics[width=\linewidth]{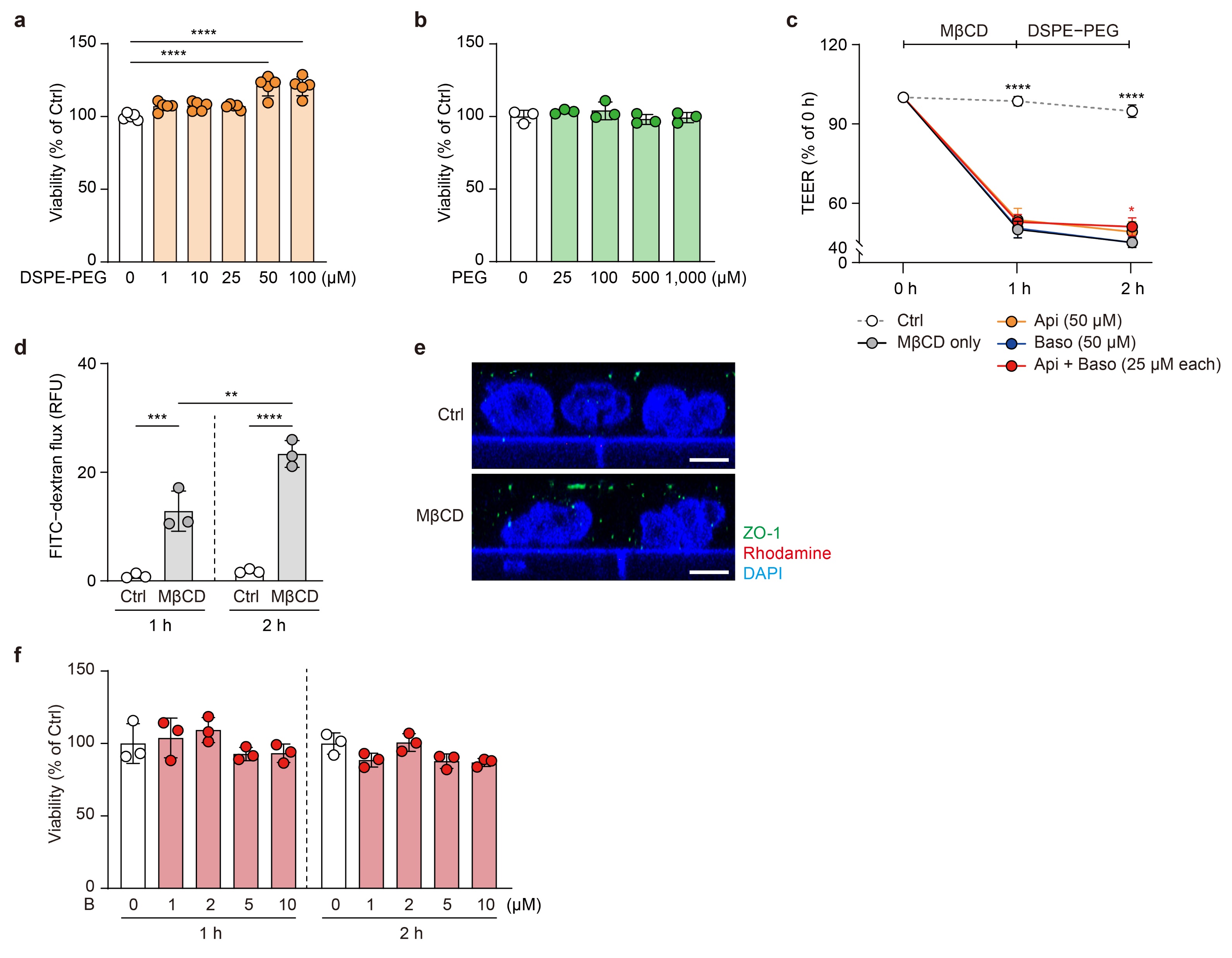}
  \caption[Optimization and control experiments]{%
    \textbf{Optimization and control experiments for PEG- and DNA-mediated
    barrier restoration.}
    \textbf{a}, Cell viability across tested
    1,2-distearoyl-\textit{sn}-glycero-3-phosphoethanolamine-poly(ethylene
    glycol) (DSPE-PEG) concentrations.
    \textbf{b}, Cell viability across tested free PEG concentrations.
    \textbf{c}, Initial comparison of DSPE-PEG tethering geometries and doses
    (apical-only, basolateral-only, and combined apical--basolateral treatment).
    Color-matched asterisks indicate comparisons with the M$\beta$CD-only
    condition.
    \textbf{d}, Pilot comparison of 1- and 2-h fluorescein isothiocyanate
    (FITC)--dextran exposure to optimize incubation time for paracellular flux
    measurements.
    \textbf{e}, Confocal Z-stack XZ reconstructions of untreated and
    M$\beta$CD-treated monolayers showing ZO-1 (green), rhodamine (red), nuclei
    counterstained with 4$'$,6-diamidino-2-phenylindole (blue). Scale bars,
    5 $\mu$m.
    \textbf{f}, Cell viability across tested cholesterol-modified DNA
    oligonucleotide concentrations. Viability in \textbf{a}, \textbf{b}, and
    \textbf{f} was assessed using a WST-1 assay (DoGenBio, Seoul, Korea) at
    450 nm. $^{*}p < 0.05$, $^{**}p < 0.01$, $^{***}p < 0.001$, and
    $^{****}p < 0.0001$, determined by one-way ANOVA with Dunnett's multiple
    comparisons test (\textbf{a}), two-way repeated-measures ANOVA with
    Dunnett's multiple comparisons test (\textbf{c}), or one-way ANOVA with
    Bonferroni's multiple comparisons test (\textbf{d}).%
  }
  \label{sfig:controls}
\end{figure}

\clearpage

\subsection[Supplementary Appendix 1. Original nanobridge proposal]{Supplementary
  Appendix 1. Original Abelio-generated cholesterol-DNA nanobridge proposal}
\label{sapp:proposal}

The following pages reproduce the original Abelio-generated cholesterol-DNA
nanobridge proposal selected for prospective experimental validation, without
modification.

% One proposal page per sheet, cut out of the 2x2 montages -- see
% \proposalpage in main.tex. The appendix keeps the document's own margins,
% so these pages are exactly as wide as every other page; the images run to
% \linewidth. Page 1 shares this sheet with the heading above, so its height
% is capped at what is left -- at 0.83\textheight it still comes out 99% as
% wide as the others, which reads as the same size.
\proposalpage[0.83\textheight]{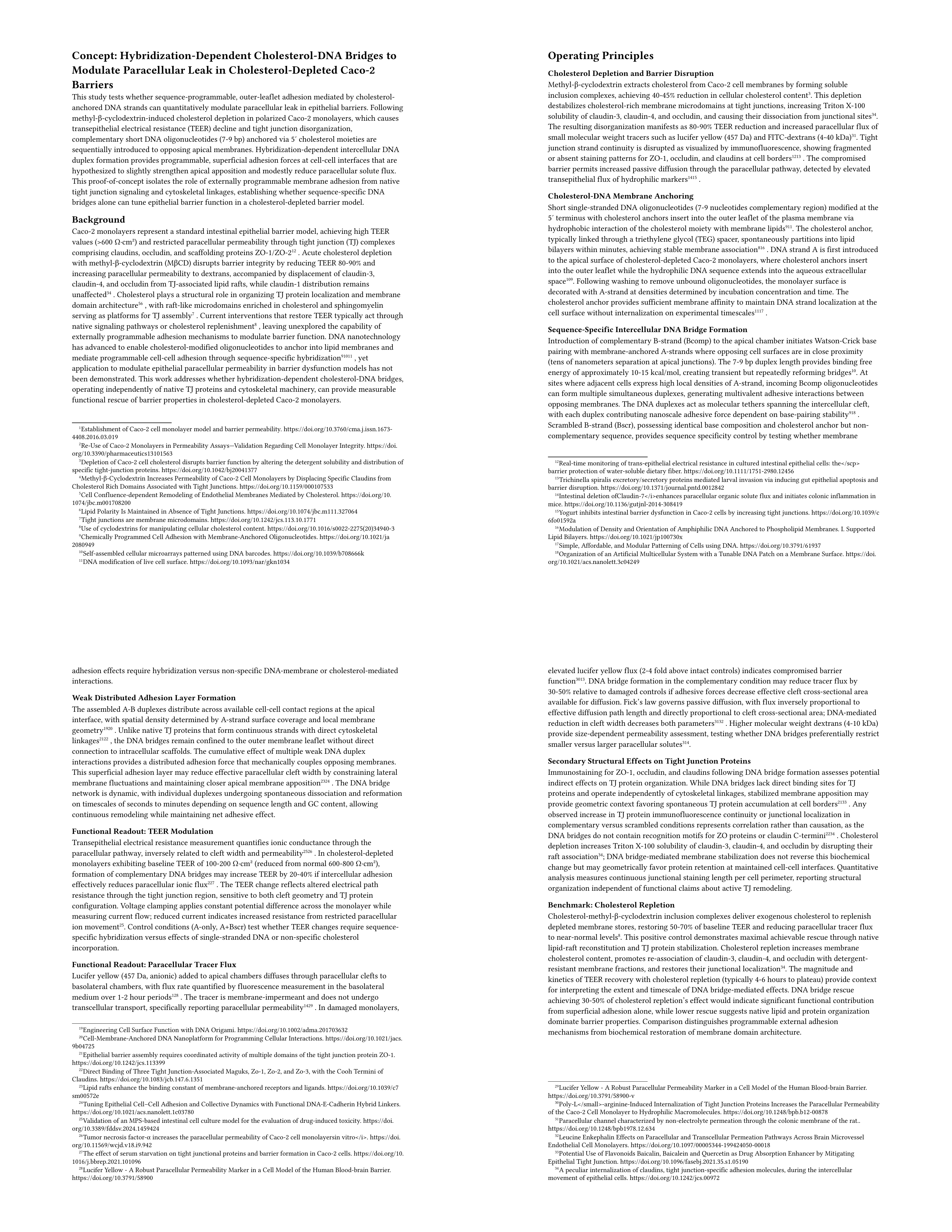}{0 396 306 792}

\clearpage
\proposalpage{resources/figures/appendix1_p1.png}{306 396 612 792}

\clearpage
\proposalpage{resources/figures/appendix1_p1.png}{0 0 306 396}

\clearpage
\proposalpage{resources/figures/appendix1_p1.png}{306 0 612 396}

\clearpage
\proposalpage{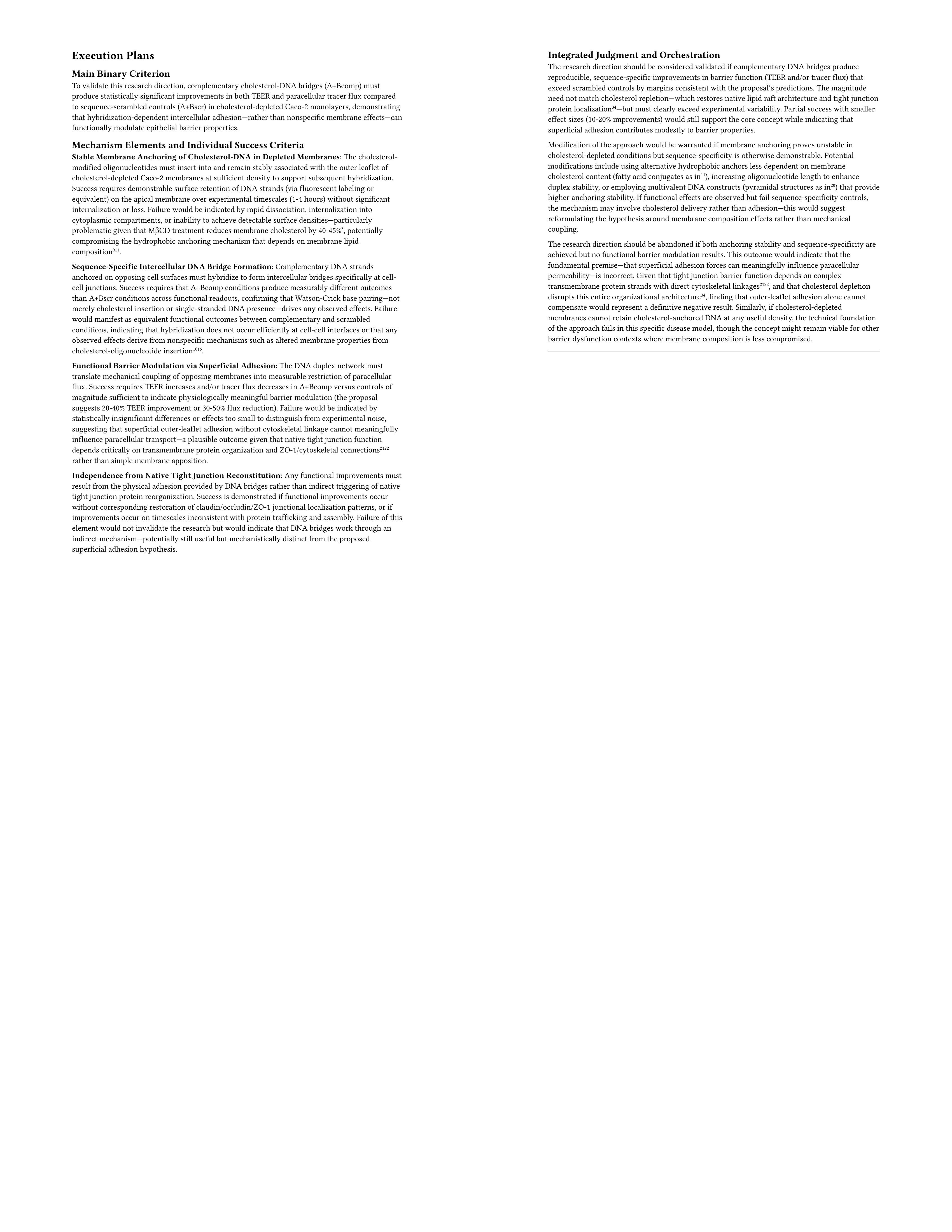}{0 396 306 792}

\clearpage
\proposalpage{resources/figures/appendix1_p2.png}{306 396 612 792}

\clearpage

\end{document}